\documentclass[11pt]{article}
\usepackage{hyperref}

\usepackage[final]{acl}

\usepackage{times}
\usepackage{latexsym}
\usepackage{fontawesome5}
\usepackage[T1]{fontenc}

\usepackage[utf8]{inputenc}

\usepackage{amstext}
\usepackage{microtype}

\usepackage{inconsolata}
\usepackage{etoc}
\usepackage{subcaption}
\usepackage{graphicx}
\usepackage{url}
\usepackage{adjustbox}
\usepackage[ruled,vlined]{algorithm2e}
\usepackage{amsmath}
\usepackage{amssymb}
\usepackage{longtable}
\usepackage{booktabs}
\usepackage{makecell}      
\usepackage{multirow}      
\usepackage{hhline}        
\usepackage{colortbl}
\usepackage{array}
\usepackage{xcolor}
\definecolor{lightgray}{gray}{0.9}
\usepackage[most]{tcolorbox}  
\usepackage[utf8]{inputenc}
\usepackage{textcomp}
\usepackage{outline}
\newtcolorbox{takeawaybox}{
  colback=blue!5!white,  
  colframe=blue!75!black, 
  fonttitle=\bfseries,
  title=Takeaways 
}

\usepackage[colorinlistoftodos]{todonotes}

\definecolor{titlebg}{gray}{0.85}
\definecolor{boxbg}{gray}{0.95}

\newcommand{\icon}[2]{%
  \raisebox{-0.3em}{\includegraphics[height=1.3em]{img/icons/#1.png}}%
}
\definecolor{myGreen}{RGB}{68, 166, 142}
\definecolor{myRed}{RGB}{248, 106, 106}
\definecolor{myPurple}{RGB}{182, 174, 234}

\title{What Makes an LLM a Good Optimizer? \\A Trajectory Analysis of LLM-Guided Evolutionary Search}

\definecolor{DodgerBlue}{RGB}{14, 88, 183} 

\author{
\textbf{Xinhao Zhang}, 
\textbf{Xi Chen},
\textbf{François Portet},
\textbf{Maxime Peyrard}
\\
\textsuperscript{}Univ. Grenoble Alpes, CNRS, Grenoble INP, LIG, 38000 Grenoble, France
\\
\texttt{\{}%
\href{mailto:Xinhao.Zhang@univ-grenoble-alpes.fr}{\color{black}\texttt{{xinhao.zhang}},} %
\href{mailto:maxime.peyrard@univ-grenoble-alpes.fr}{\color{black}\texttt{{maxime.peyrard}}}%
\texttt{\}@univ-grenoble-alpes.fr}
\\
\\
Project Website: \href{https://xinhao-zhang.github.io/traj_evo_search/}{\color{DodgerBlue}github.io/traj\_evo\_search}}

\begin{document}
\maketitle

\begin{abstract}
Recent work has demonstrated the promise of orchestrating large language models (LLMs) within evolutionary and agentic optimization systems. However, the mechanisms driving these optimization gains remain poorly understood. In this work, we present a large-scale study of LLM-guided evolutionary search, collecting optimization trajectories for 15 LLMs across 8 tasks. Although zero-shot problem-solving ability correlates with final optimization outcomes, it explains only part of the variance: models with similar initial capability often induce dramatically different search trajectories and outcomes. By analyzing these trajectories, we find that strong LLM optimizers behave as local refiners, producing frequent incremental improvements while progressively localizing the search in semantic space. Conversely, weaker optimizers exhibit large semantic drift, with sporadic breakthroughs followed by stagnation. Notably, various measures of solution novelty do not predict final performance; novelty is beneficial only when the search remains sufficiently localized around high-performing regions of the solution space. Our results highlight the importance of trajectory analysis for understanding and improving LLM-based optimization systems and provide actionable insights for their design and training.
\end{abstract}

\section{Introduction}

%
%
%
%

Large language models (LLMs) are increasingly deployed as search operators in iterative optimization systems \citep{lehman2023evolution,peyrard2024agentic}. Across diverse domains, such as prompt optimization \citep{agrawal2026gepareflectivepromptevolution,guo2023connecting,fernando2023promptbreeder} and scientific discovery \citep{romera2024mathematical,ellenberg2025generative,novikov2025alphaevolve,gottweis2025towards}, LLMs are embedded into evolutionary or agentic loops as black-box optimizers where they repeatedly propose candidate solutions, receive feedback, and refine solutions iteratively. 
While such LLM-guided evolutionary workflows have been shown to deliver substantial empirical gains, the mechanisms underlying these improvements remain poorly understood. In particular, even under strictly controlled optimization loops, selection rules, and evaluation functions, different LLMs exhibit vastly different optimization trajectories and final performances. This observation motivates the central question of this work: \emph{\textbf{what explains such large model-to-model differences in optimization performance?}} Are these differences primarily a reflection of base model capability, or do they arise from more subtle differences in the exploration--exploitation dynamics induced by the models?
%
%
\begin{figure}[t]  
    \centering
    \includegraphics[width=\columnwidth]{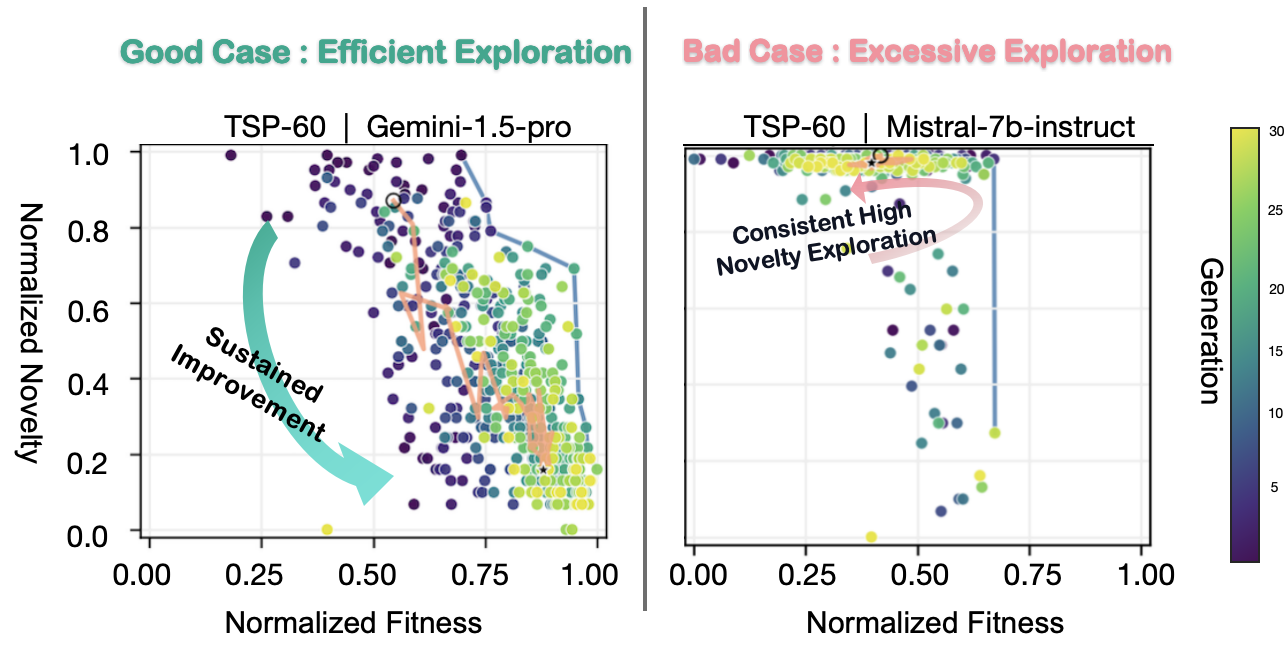}
    \caption{\textbf{Different optimization trajectories for two LLMs with similar zero-shot performance on TSP-60.} Each point represents a candidate solution, colored by generation. Gemini-1.5-Pro (left) displays sustained fitness improvement and progressive localization. Mistral-7B-Instruct (right) maintains high novelty but fails to exploit it into fitness gains.}
    \label{fig:novelty_compare}
\end{figure}

To address these questions, we conduct a large-scale study of LLM-based evolutionary optimization, collecting optimization trajectories for 15 LLMs across 4 task families (8 tasks), resulting in 72K analyzed candidate solutions. 
As expected, zero-shot performance correlates positively with final optimization outcomes. However, this relationship explains partially the variance: models with nearly identical zero-shot performance can diverge quickly after evolution, following qualitatively distinct optimization trajectories (see Figure~\ref{fig:novelty_compare}).
To explain this residual gap, we turn to the classical exploration--exploitation trade-off underlying evolutionary algorithms~\citep{sun2018balancing}. In LLM-driven metaheuristics, the mutation operator is no longer fully random but is strongly shaped by the LLM’s prior toward producing improved solutions conditioned on parent candidates and their fitness feedback. Consequently, exploration is more constrained than in classical, non-LLM-driven evolution. Under this view, optimization processes should benefit primarily from higher novelty and diversity, which would expose the system to a broader range of potentially useful regions in the search space.
Surprisingly, our results contradict this intuition. We find that the most successful trajectories are not those exhibiting high novelty, but rather those characterized by \emph{frequent and sustained breakthroughs}, where a breakthrough represents an incremental improvement of the best-so-far fitness. In other words, what distinguishes strong optimization runs is more of the ability to reliably produce incremental improvements repeatedly. This is different from typical behavior observed in meta-heuristics, where large breakthroughs occur at rare intervals, followed by long plateaus or small refinements~\citep{mitchell1999evolutionary}. This interpretation is also supported by our perturbation experiments, where we directly manipulate the refinement behaviour of the search trajectory through model mixing, leading to predictable changes in optimization performance.

To further elucidate when and why breakthroughs happen, we analyze the \textbf{geometry of optimization trajectories in semantic space}. By embedding candidate solutions and characterizing within-generation distributions with entropy-based and dispersion measures, we reveal a clear distinction between strong and weak optimizers. Effective LLM operators progressively \emph{localize} their search around high-performing regions of semantic space, whereas weaker optimizers continue to diffuse and drift across distant regions. A generation-level mixed-effects analysis further uncovers an interesting interaction between novelty and semantic dispersion: novelty increases the probability of breakthroughs \emph{only when} the search remains sufficiently localized. Outside this regime, novelty is largely unproductive.
%

\textbf{Contributions.} We make the following contributions: 
\textbf{(i)} We conduct a large-scale, controlled study of LLM-based evolutionary optimization and release the resulting optimization trajectories. 
\textbf{(ii)} We show that differences in optimization performance between LLMs are only partially explained by zero-shot capability, unveiling a distinct notion of \emph{optimizable ability}. 
\textbf{(iii)} We identify effective LLM optimizers as \emph{local refiners}, whose trajectories progressively localize in semantic space and yield frequent incremental breakthroughs, and we support this mechanism through perturbation experiments that highlight the role of refinement behavior.
\textbf{(iv)} We demonstrate that novelty is not inherently beneficial; its utility is conditional on the geometric regime of search, and it becomes productive only when search remains localized.
\textbf{(v)} We derive practical implications for model selection and for learning better search operators, showing that smaller or cheaper models can outperform stronger base models when they exhibit more reliable refinement behavior.

More broadly, our semantic trajectory analysis offers a reusable framework for studying LLM-driven optimization processes. Our findings suggest that, rather than solely pursuing general-purpose capability, future work may benefit from understanding, controlling, training models as effective search operators~\citep{surina2025algorithm} emphasizing local refinement and error correction.

\section{Related Work}

\paragraph{Evolutionary Computation with LLMs}  
LLMs are increasingly integrated into evolutionary computation frameworks~\citep{yang2023large,wu2024evolutionary,brahmachary2024largelanguagemodelbasedevolutionary,tao2024survey}, revitalizing meta-heuristic optimization. Unlike classical approaches that rely on stochastic operators to explore optimization landscapes~\citep{holland1992adaptation}, LLM-assisted methods instantiate variation operators through the semantic priors of LLMs~\citep{josifoski2023flows,peyrard2024agentic,gao2025survey,fang2025comprehensive}. These frameworks have demonstrated strong empirical performance across domains including combinatorial optimization~\citep{yang2025heuragenixleveragingllmssolving,yu2026largelanguagemodeldrivenfullcomponent} and scientific discovery~\citep{yang2024moose,macknight2025pretrainedknowledgeelevateslarge,chen2026molevolvellmguidedevolutionarysearch,abhyankar2026llemaevolutionarysearchllms,zhou2024hypothesis}. Recent work has further extended these approaches to algorithm discovery in open-ended scenario~\citep{lu2024aiscientistfullyautomated,gottweis2025towards,qu2026coralautonomousmultiagentevolution}, as well as more advanced settings incorporating co-evolution or meta-reflection to mitigate limited global search perspectives~\citep{liu2024evolution,ye2024reevo}. Our work complements these efforts by analyzing the search behavior and optimization trajectories induced by LLMs. Trajectory analyses can also inform the design of agentic systems~\citep{lee2026tmapredteamingllmagents,zhao2026largelanguagemodelpoweredevolutionary,zhao2025trajevodesigningtrajectoryprediction,lin2025seagentselfevolutiontrajectoryoptimization}. Related to our work are behaviour-space studies~\citep{vanstein2025behaviourspaceanalysisllmdriven} and LAS landscape analyses~\citep{liu2025fitnesslandscapelargelanguage}, which similarly associate effective optimization with sustained improvements and increased exploitation. We extend these findings with a unified cross-model, cross-task analysis using semantic entropy measures.

\paragraph{Evaluating LLMs as Search Operators}
The evaluation paradigm for LLMs has evolved accordingly. 
Early optimization benchmarks relied on single-pass prompting~\citep{fan2024nphardeval,duchnowski2025ehop}. 
More recent work evaluates LLMs within iterative or evolutionary search loops, treating them as search operators guided by external feedback~\citep{li2025opt,huang2024exploring,ouyang2025kernelbench,shojaee2024llm,shojaee2025llm}. 
In this setting, LLMs are no longer assessed as one-shot solvers but as \emph{semantic search operators}, whose preferences and biases matter~\citep{zhou-etal-2026-matters}. 
While these benchmarks demonstrate strong end-to-end performance, evaluation remains largely outcome-centric. 
Our results show that base model capability and operator effectiveness are distinct skills, with direct implications for model selection and motivating work on learning specialized search operators, such as~\citet{brahmachary2024largelanguagemodelbasedevolutionary} and EvoTune~\citep{surina2025algorithm}.

\begin{figure*}
    \centering
    \includegraphics[width=0.98\textwidth]{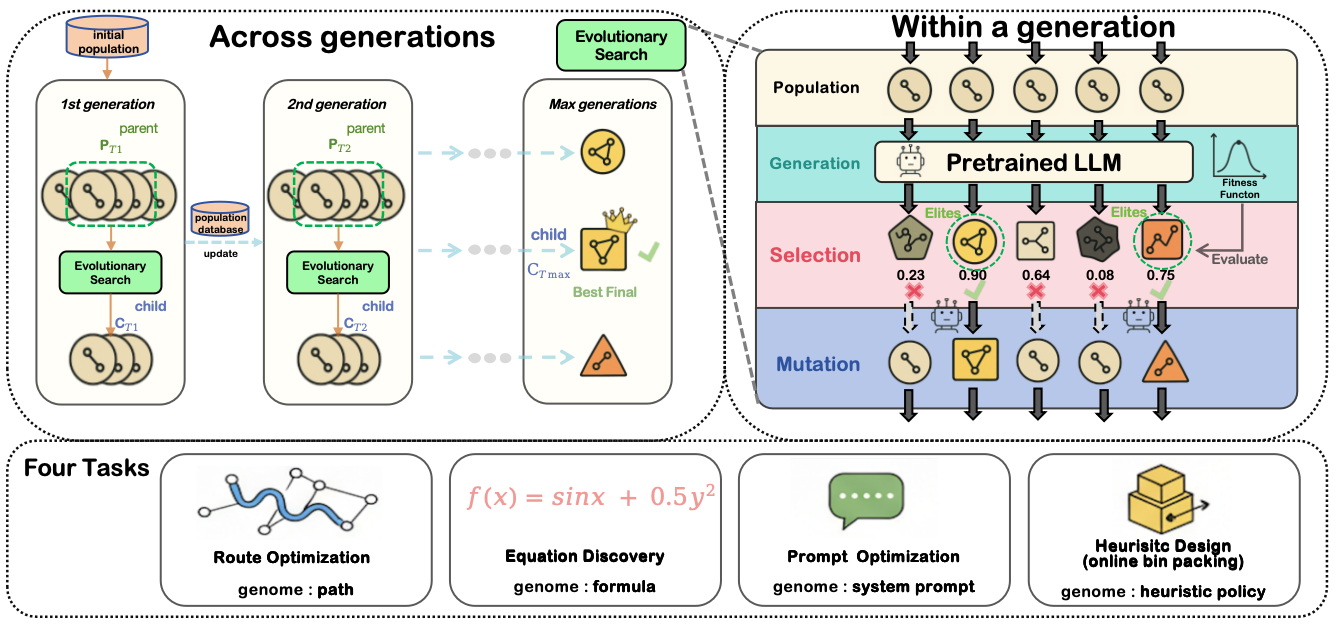}  
    \caption{\textbf{Overview of the LLM-driven evolutionary search framework and tasks.} Left: the evolutionary process across generations. Right: the within-generation loop—population initialization, LLM-guided mutation, fitness evaluation, and selection. Bottom: the four tasks and their corresponding genome representations.}  
    \label{fig:eval_framework}
\end{figure*}

\section{Methodology}\label{sec:protocol}

Following \citet{novikov2025alphaevolve}, we implement a lightweight evolutionary search loop where LLMs act as semantic variation operators, iteratively generating candidate solutions in order to optimize task-specific fitness.

%


\smallskip      
\noindent
\textbf{Population Initialization.} For each task, we construct an initial population $\mathcal{P}_0$ consisting of valid genomes and their corresponding fitness values $(g, f_T(g))$. This initial population is fixed and shared across all models for the same task.


\smallskip
\noindent
\textbf{Fitness Evaluation.} Each genome is evaluated by a task-specific fitness function $f_T(\cdot)$. Invalid or unparsable outputs are assigned zero fitness. 


\smallskip      
\noindent
\textbf{Selection (Top-$q$ Weighted).} At generation $t$, we form an elite subset $\mathcal{E}_t = \operatorname{Top}_{\lceil qN \rceil}(\mathcal{P}_t)$ with $q$ fixed to $0.2$. Parents are sampled from $\mathcal{E}_t$ with probability proportional to their fitness:
$
\Pr(x \mid \mathcal{E}_t) \propto f_T(x)
$.


\smallskip      
\noindent
\textbf{Mutation.} Selected parent genomes are provided as the context of prompts to the LLM, which generates a set of offspring genomes $\mathcal{C}_t$ conditioned on the task and parent structure.


\smallskip
\noindent
\textbf{Population Pool Update.} Generated offspring are deduplicated and merged into the population pool. If the pool size exceeds $N$, only the top-$N$ genomes ranked by fitness are retained. The best-so-far fitness is updated as $f^\star_t = \max_{x \in \mathcal{P}_t} f_T(x)$.



\subsection{Tasks \& Genome Representations} 
Our evaluation includes tasks across four domains, spanning combinatorial, linguistic, symbolic, and algorithmic optimization where previous work has shown benefits from LLM-guided search. 

\smallskip
\noindent
\adjustbox{valign=b}{\includegraphics[height=1.3em]{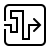}}~
\textbf{Route Optimization} \quad The Traveling Salesman Problem (TSP) is a classical route optimization task. Prior work shows that while LLMs struggle to produce high-quality tours in a single pass \citep{fan2024nphardeval}, evolutionary search can largely improve performance \citep{huang2024exploring}. For each optimization run, one randomly generated distance matrix is given and the output must be a valid tour as a permutation of city indices. We evaluate two TSP variants with 30 and 60 cities, respectively. Each genome represents a permutation of the tour, and the fitness function is the inverse total distance.



\smallskip
\noindent
\adjustbox{valign=b}{\includegraphics[height=1.3em]{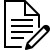}}~
\textbf{Prompt Optimization} \quad It aims to automatically improve prompts quality. The genome is a textual instruction that conditions a frozen LLM (\texttt{gpt-4o-mini}). Evolutionary approaches have been shown effective in this setting \citep{guo2023connecting,fernando2023promptbreeder}. We evaluate on dialogue summarization (SAMSum \citep{gliwa2019samsum}) and text simplification (ASSET \citep{alva2020asset}). 
Fitness is computed as the average generation quality on a held-out 25\% validation subset, using ROUGE-L for SAMSum and SARI for ASSET, following \citet{guo2023connecting}.


\smallskip
\noindent
\adjustbox{valign=b}{\includegraphics[height=1.2em]{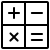}}~
\textbf{Equation Discovery} \quad Symbolic regression aims to discover concise mathematical expressions that fit observed input–output pairs \citep{grayeli2024symbolic}. This domain is well-suited for LLM-guided evolutionary search which combines prior scientific knowledge with iterative refinement \citep{shojaee2024llm,shojaee2025llm}. We adopt two nonlinear oscillation benchmarks from \citet{shojaee2024llm}: Oscillator~1 (three variables) and Oscillator~2 (four variables). Each solution encodes a candidate symbolic expression executable as $f(x)$. Fitness is measured as $f_T(\text{expr}) = 1 - \operatorname{norm}(\mathrm{MSE}(\hat{y}, y))$, where $\hat{y}$ denotes model predictions, and $\operatorname{norm}$ is the min--max normalization computed over all candidate solutions for the same task instance.


\smallskip
\noindent
\adjustbox{valign=b}{\includegraphics[height=1.3em]{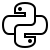}}~
\textbf{Heuristic Design}\quad Heuristic design for combinatorial optimization aims to evolve executable programs rather than direct solutions. This paradigm has been successfully explored in recent work such as FunSearch and EoH \citep{romera2024mathematical,liu2024evolution}. We focus on the online bin packing problem. Each genome encodes a heuristic policy in the form of a priority function that determines item placement. We evaluate on two datasets: OR3 (20 instances, 500 items each) and Weibull (5 instances, 5,000 items each), representing synthetic and real-world-like distributions. Fitness is the inverse number of bins used.

\subsection{Novelty Computation}\label{Sec:novelty_computation}

\paragraph{Task-agnostic novelty.} Besides fitness, we further quantify semantic diversity along the trajectory. We defined the novelty of a solution $a$ with respect to a set of all previous solutions $\mathcal{A}^{\mathrm{prior}}$ including the initial parents under a task-specific semantic distance metric $D_T$: $nov(a, \mathcal{A}^{\mathrm{prior}}) = \min_{b \in \mathcal{A}^{\mathrm{prior}}} D_T(a, b)$.  Novelty is normalized at subtask-level to ensure comparability.



\paragraph{Task-specific semantic distance.} For TSP, we use an edge-set distance invariant to rotation and starting city. For prompt optimization, we compute cosine distance in a fixed embedding space using OpenAI's \texttt{text-embedding-ada-002}. For equation discovery and heuristic design, we adopt a functional behavior distance measured over a fixed input grid to capture divergence in output behavior.

\subsection{Evolution Scale and Parameters}\label{evo_param}

Our study involves 15 LLMs, with 30 generations conducted for each (model, task) pair. In each generation, the population produces 10 offspring, each corresponding to a model call. Every model–task pair is repeated twice with the same initial population, thereby totaling over 72,000 API calls. All evolutions are conducted using a default temperature of 0.7. The total cost of running experiments is estimated to be around \$500.\footnote{All trajectory data is available at \url{https://huggingface.co/datasets/LivevreXH/evo_llm_trajectories}.}

%


The selected LLMs span six model families:
~\icon{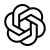}~ OpenAI (GPT-4o \citep{hurst2024gpt}, GPT-4o-mini, GPT-3.5-turbo), Google's~\icon{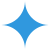}~ Gemini (Gemini-1.5-flash, Gemini-1.5-pro \citep{team2024gemini}), Google's~\icon{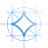}~ Gemma-3n-4b \citep{team2025gemma}, Meta's~\icon{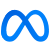}~ Llama (llama-3.1-70b-instruct, llama-3.1-8b-instruct, llama-3.2-1b-instruct, llama-3.2-3b-instruct \citep{grattafiori2024llama}), Deepseek AI's~\icon{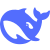}~ Deepseek-V3 \citep{liu2024deepseek}, MistralAI's~\icon{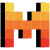} ~ Mistral (Mistral-7b-instruct \citep{jiang2023mistral7b}, Mistral-24b-instruct, Mistral-large, Magistral-small \citep{rastogi2025magistral}). Additional experimental details are provided in Appendix~\ref{evolution_protocol_details}. 


\section{Results and Analysis}

\paragraph{The Optimization Gap.} We evaluate each LLM under a fixed evolutionary budget and conditions. Performance is measured by the best fitness at the end of evolution. Across task families, we observe a pronounced \textbf{optimization gap} between models. Concretely, under identical conditions, different LLMs lead to different optimization outcomes (see Table~\ref{tab:results_rq1}). Strong early performance does not reliably predict long-horizon outcomes. For instance, Deepseek-V3 performs the best in the first generation yet fail to achieve the largest gains over time.  This suggests that LLMs differ not only in solution quality, but in the search process. In the following section, we progressively rule out alternative explanations, i.e. base capability, novelty, and identify a consistent mechanism for successful optimization.



\subsection{Base Model Capability}\label{sec:base_model_analysis}


We first hypothesize that the gap may simply stem from base model capability, specifically the model's intrinsic task-specific problem-solving ability in zero-shot settings. Since all tasks are optimization-oriented, we define \textbf{zero-shot performance} as the best fitness achieved via temperature-swept \emph{best-of-$N$} sampling: for each model--task pair, we generate candidates across six temperatures ($T \in \{0.0, 0.2, 0.4, 0.6, 0.8, 1.0\}$), with two samples per temperature, and report the best fitness among them (See Appendix~\ref{evolution_protocol_details} for more details).

\begin{figure}[tbp]  
    \centering
    \includegraphics[width=0.89\columnwidth]{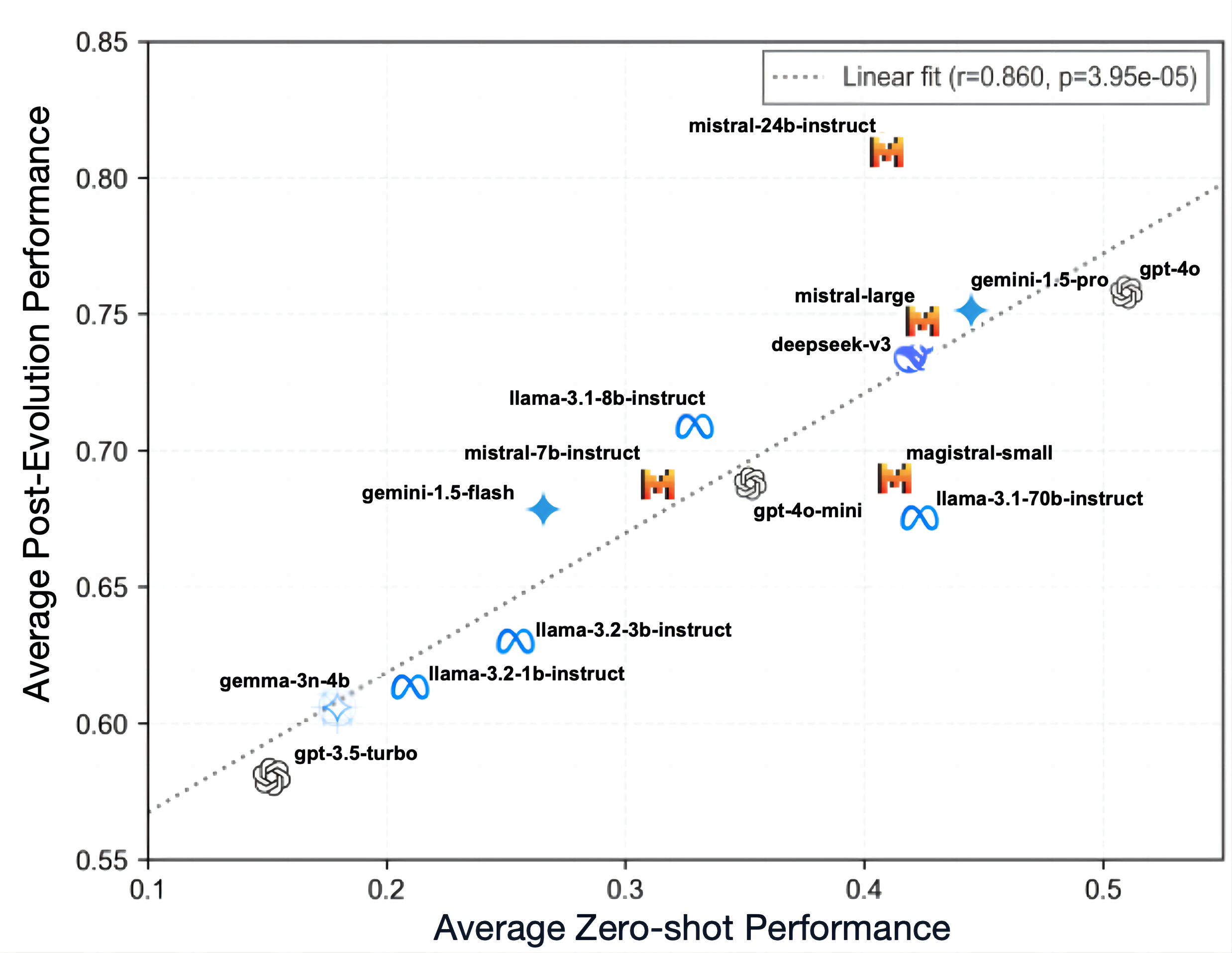}
    \caption{Scatter plot between zero-shot performance and final optimized performance across models.}
    \label{fig:zero_compare}
\end{figure}

As revealed in Figure~\ref{fig:zero_compare}, zero-shot performance is strongly correlated with post-optimization performance when aggregated across tasks. Similar trends are observed at the sub-task level (Appendix Figure~\ref{fig:task_zero_perf_best}), where higher zero-shot scores generally correspond to better final outcomes, only except for equation discovery tasks. This confirms that base capability is a strong predictor of optimization potential, yet insufficient to fully explain long-horizon optimization success. Models with nearly identical zero-shot performance can diverge substantially after evolution. For instance, around an average zero-shot score of 0.4 in Figure~\ref{fig:zero_compare}, multiple models cluster tightly along the zero-shot axis yet spread widely in their best final performance. This residual variance persists across tasks (See Figure~\ref{fig:task_zero_perf_best}).

\subsection{Trajectory-Level Analysis}

Since being a strong one-shot problem solver does not necessarily imply being an effective evolutionary search operator, the differences might lie in the search process. We therefore investigate trajectory-level properties of evolutionary search.

\subsubsection{Novelty vs. Breakthrough Dynamics}\label{sec:novelty_trap}

 In classical evolutionary algorithms, novelty is treated as a proxy for exploration. However, this equivalence becomes problematic in LLM-guided evolution. Different from \textit{blind} stochastic mutation operators, LLMs generate offspring by conditioning on parent solutions and task context aiming to produce the best solution. Novelty in this setting does not arise from random exploration, but from semantic variation within the LLM's output.

\begin{figure}[tbp]  
    \centering
    \includegraphics[width=0.99\columnwidth]{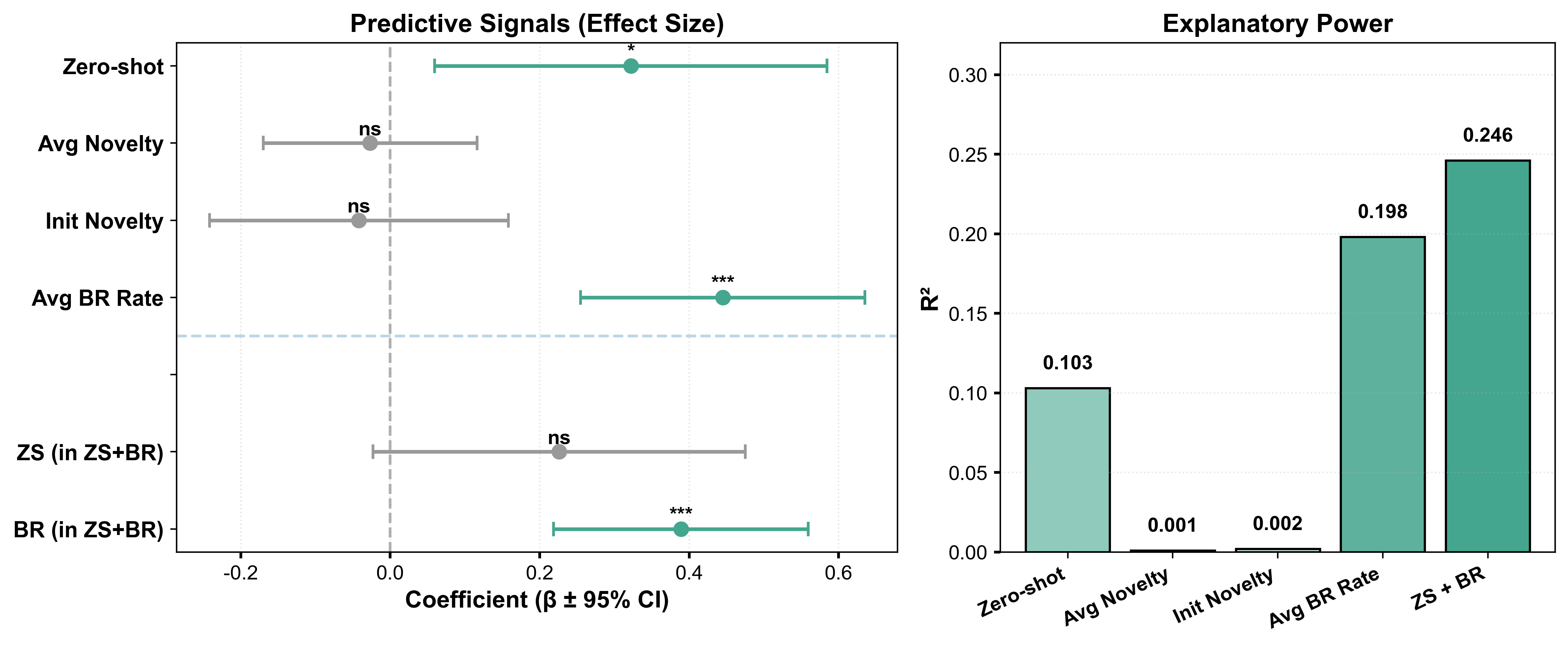}
\caption{\textbf{OLS regression results across different trajectory descriptors.} (Left) Standardized coefficients. (Right) Explanatory power. $^{***}p<0.001$, $^{*}p<0.05$, $ns$ means non-significant p-values. Novelty-based predictors are not significant, whereas breakthrough rate (BR) strongly predicts performance and improves fit beyond zero-shot capability (ZS). }
    \label{fig:nov_regre}
\end{figure}

\paragraph{More Novelty Doesn't Yield Better Optimization.} A natural hypothesis is therefore that models generating more novel solutions should explore the search space more effectively and achieve better optimization outcomes. We test this hypothesis in Figure~\ref{fig:nov_regre}, which summarizes both effect sizes and explanatory power across different trajectory-level descriptors. Astonishingly, novelty-based measures, including both average novelty and initial novelty (in the first generation), exhibit coefficients close to zero and are not statistically significant. Moreover, their explanatory power is negligible, meaning that increasing diversity alone does not contribute to improved optimization performance.

\paragraph{Breakthrough Rate Strongly Predicts Optimization Performance.}  We define a \textbf{breakthrough} as a best-so-far improvement event, e.g., an offspring generation in which the current solution exceeds the best fitness solution in all previous generations. We quantify each optimization trajectory's tendency to produce breakthroughs by its \textbf{breakthrough rate}. The breakthrough rate is also averaged per pair of models and tasks. As shown in Figure~\ref{fig:nov_regre} (left), the breakthrough rate has the largest positive coefficient among all predictors. This is further reflected in its explanatory power (right), where breakthrough rate alone explains around two times more variance than zero-shot capability. Beyond this, when combining breakthrough rate with zero-shot performance, the overall explanatory power increases further, while the coefficient of zero-shot performance decreases. This indicates that part of the predictive power of base capability is mediated through the ability to generate consistent improvements during search. 

These results show that good optimization trajectories tend to frequently produce small improvements instead of big and rare breakthroughs followed by long plateaus, as it often happens in evolutionary search~\citep{mitchell1999evolutionary}.

\subsubsection{Semantic Geometry}\label{sec:semantic_analysis}

In LLM-guided evolution, mutation operators are black-boxes, making it difficult to directly interpret how search progresses. To understand why some trajectories yield more breakthroughs than others, we instead examine the geometry of the search process by analyzing how solutions are distributed in semantic space over time.

 \begin{figure*}[t]
    \centering
    
    \begin{minipage}[c]{0.48\textwidth} 
        \begin{subfigure}{\linewidth}
            \centering
            \includegraphics[width=0.95\linewidth,height=3.8cm, keepaspectratio=false]{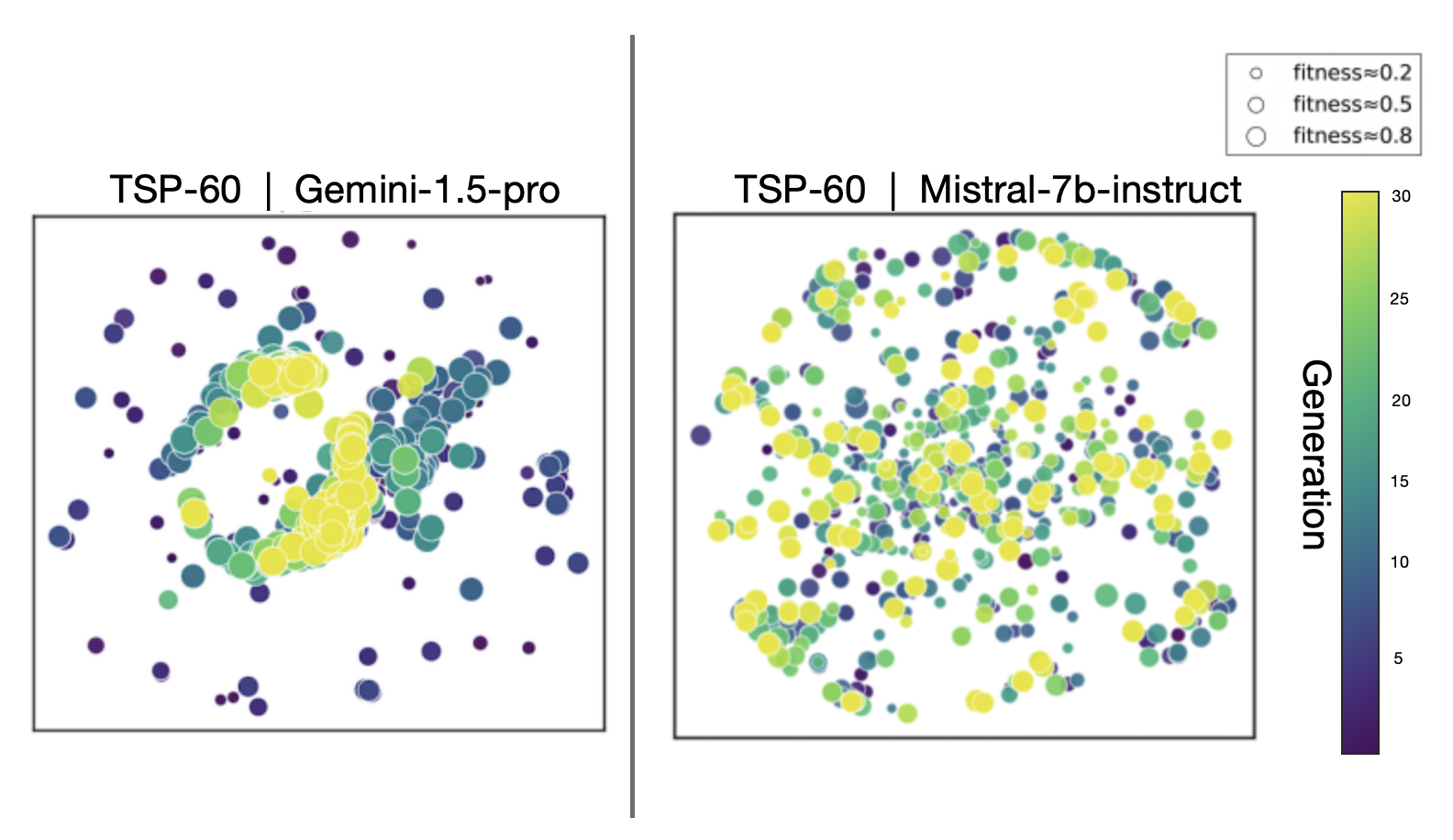} 
            \caption{Search Topology Visualization (MDS)}
            \label{fig:mds_topology}
        \end{subfigure}
        
        \par\medskip 
        
        \begin{subfigure}{\linewidth}
            \centering
            \includegraphics[width=0.92\linewidth,height=3.6cm, keepaspectratio=false]{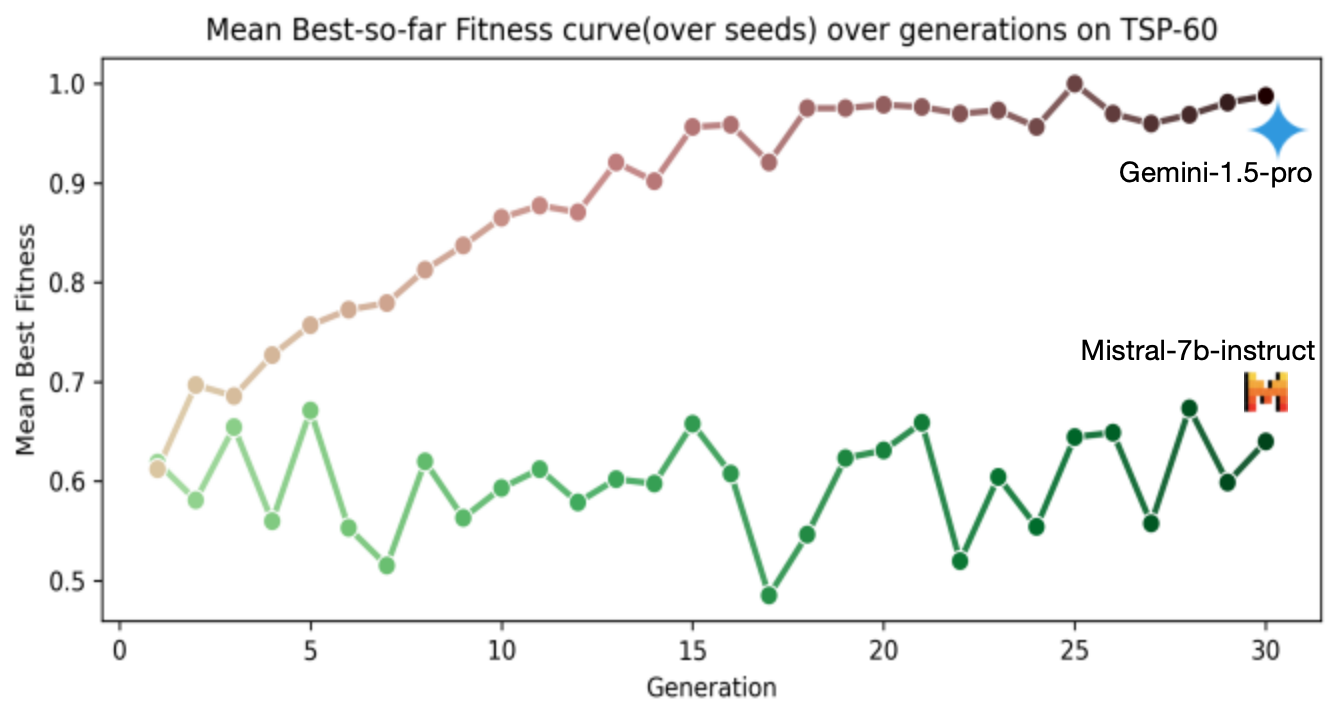}
            \caption{Best Fitness Progression}
            \label{fig:best_fitness}
        \end{subfigure}
    \end{minipage}%
    \hfill 
    \begin{minipage}[c]{0.46\textwidth}
        \centering
        
        \begin{subfigure}{\linewidth}
            \centering
            \includegraphics[width=\linewidth,height=3.65cm, keepaspectratio=false]{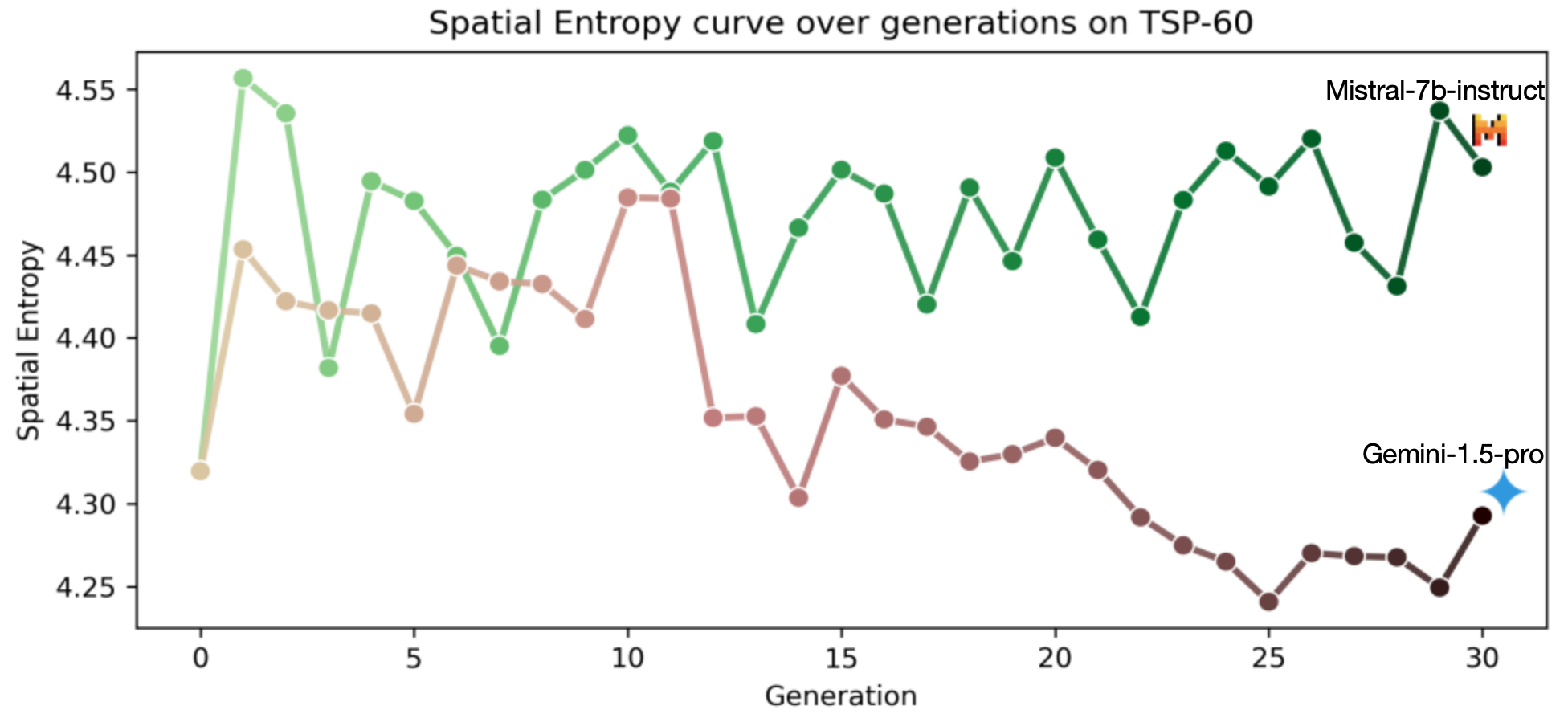}
            \caption{Spatial Entropy}
            \label{fig:spatial_entropy}
        \end{subfigure}
        
        \par\medskip 
        
        \begin{subfigure}{\linewidth}
            \centering
            \includegraphics[width=\linewidth,height=3.65cm, keepaspectratio=false]{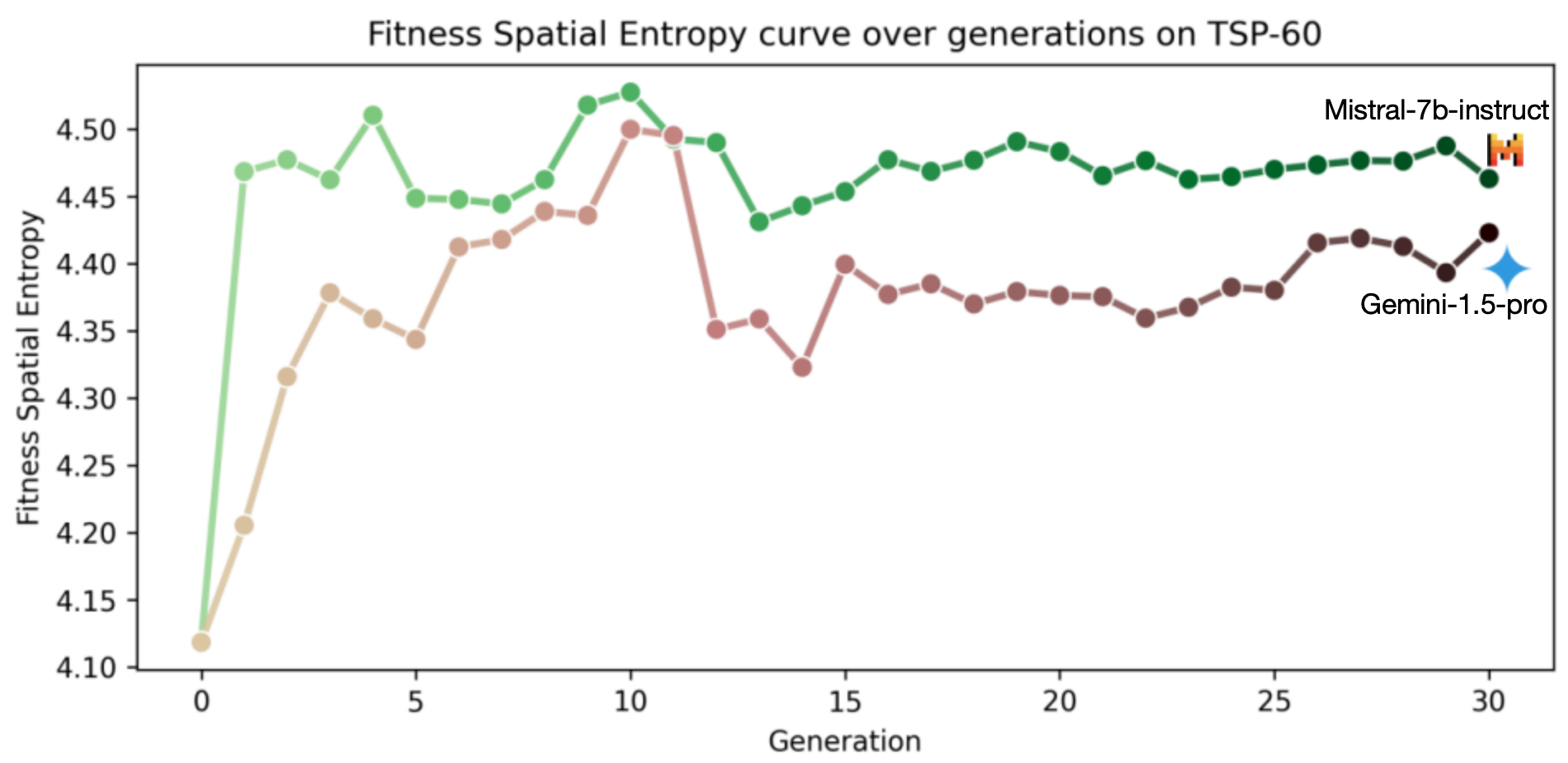}
            \caption{Fitness Spatial Entropy}
            \label{fig:fitness_entropy}
        \end{subfigure}
    \end{minipage}

    \caption{\textbf{A qualitative contrast of evolutionary search geometry analysis.} 
    (a) Visualization of the search space topology using MDS. Gemini-1.5-pro forms a convergent solution cluster (yellow). All points are projected using a shared MDS space learned from all task-specific candidates.
    (b) The Mean Best Fitness curve shows the convergence speed and quality over seeds.
    (c) Spatial Entropy quantifies the candidates' organization. 
    (d) Fitness-Spatial Entropy illustrates Gemini's solutions are high-quality and topologically concentrated.}
    \label{fig:case_topology}
\end{figure*}



We embed all candidate solutions into a task-specific shared semantic space, enabling us to analyze the within-generation distribution of candidates. 
Precisely, we measure the spatial organization of search using kernel-based entropy. 
Let $x_i \in \mathbb{R}^d$ denote the embedding of solution $i$, and $K(\cdot,\cdot)$ a Gaussian kernel. For any weighting $w_j$, we compute a local density estimate
$$
g_i = \sum_j w_j K(x_i, x_j), \quad q_i = \frac{g_i}{\sum_k g_k},
$$
and define
$$
H = -\sum_i q_i \log q_i.
$$
This framework yields two complementary views. Setting $w_j = 1$ gives \textbf{(i) spatial entropy ($H_{\text{spatial}}$)}, which measures how broadly candidates spread across semantic space. Setting $w_j = f_j$ gives \textbf{(ii) fitness spatial entropy ($H_{\text{fitness}}$)}, which measures whether high-quality solutions cluster or distribute across regions. These metrics intuitively summarize \textit{how solutions are spatially organized} within a generation as they distinguish between diffuse versus localized search globally ($H_{\text{spatial}}$), and whether high-fitness solutions concentrates or spreads across the semantic landscape ($H_{\text{fitness}}$). We also complement them with multidimensional scaling visualizations that project all solutions onto a shared two-dimensional space, with points colored by generation and scaled by fitness (All MDS plots are on our website\footnote{\url{https://xinhao-zhang.github.io/traj_evo_search/}}). Figure~\ref{fig:case_topology} illustrates a representative case. Despite similar zero-shot performance and identical initial populations, Gemini-1.5-Pro progressively localizes its search into a smaller semantic region, while Mistral-7B-Instruct continues to drift across distant regions.

\subsubsection{Generation-Level Statistical Test}\label{sec:verif_analysis}

From the preceding case study in Figure~\ref{fig:case_topology}, we hypothesize that the optimization success may specifically depend on the geometric properties of the population. We then use a generation-level mixed-effects regression analysis to examine which effects influence breakthrough events' production.

\begin{figure}[htbp]  
    \centering
    \includegraphics[width=1.1\columnwidth]{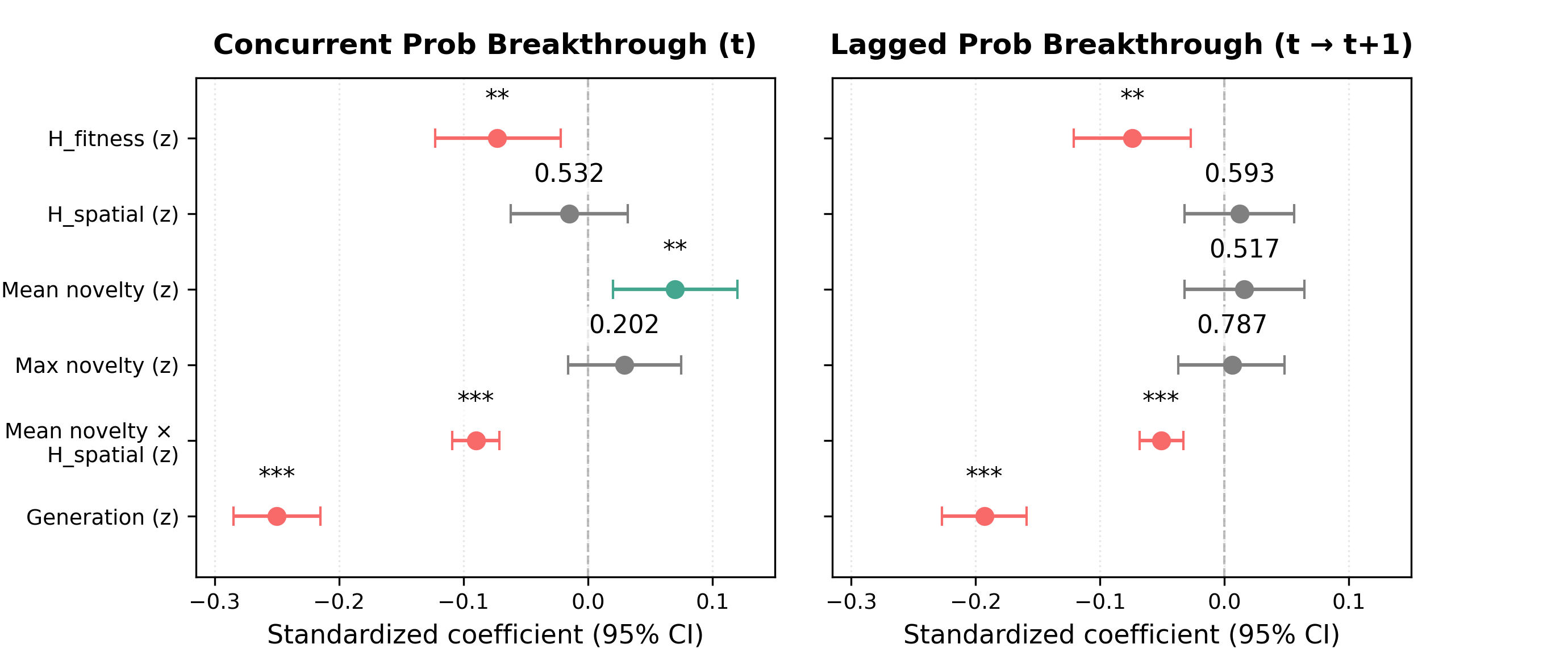}
    \caption{\textbf{Generation-level mixed-effects regression of breakthrough probabilities.} Standardized coefficients are shown for concurrent (left) and lagged (right) models, with predictors at generation $t$ explaining breakthroughs at $t$ or $t{+}1$. $^{***}p<0.001$, $^{**}p<0.01$, numeric labels report non-significant p-values.}
    \label{fig:mixed_effects_regression}
\end{figure}

We model breakthrough probability at generation $t$ as a function of population-level descriptors, including spatial entropy ($H_{\text{spatial}}$), fitness spatial entropy ($H_{\text{fitness}}$), mean and maximum novelty, generation index and their interaction for each generation. To account for repeated measurements and systematic differences across LLMs, we include model-specific random intercepts.  Results are presented in Figure~\ref{fig:mixed_effects_regression}, reporting both concurrent effects (generation $t$) and lagged effects (predicting $t+1$).
 There are several consistent patterns. First, the generation index is strongly negative in both specifications, indicating that breakthroughs occur mostly in early generations. Second, higher fitness spatial entropy is negatively associated with breakthrough probability, which counter-intuitively suggests that maintaining multiple dispersed high-quality regions would hinder breakthrough production. Third, although mean novelty is positively associated with breakthroughs within a generation, this effect is strongly conditioned on population geometry: the interaction between novelty and spatial entropy is significantly negative in both concurrent and lagged analyses. In other words, \textbf{novelty increases the likelihood of breakthroughs only when search remains sufficiently localized}. Crucially, while the effect of novelty fades under lagged prediction, the interaction effect remains significant, indicating that the productivity of novelty depends on the geometric state of the population rather than on contemporaneous correlations alone. Figure~\ref{fig:interaction_heatmap} further visualizes this interaction. Breakthrough events concentrate in regions featured by high novelty and low spatial entropy, whereas high novelty under high dispersion is usually related with low breakthrough probability.


\subsection{Operator-Level Validation}

The trajectory-level analyses above characterize \emph{what} successful optimization runs look like: strong models progressively localize in semantic space and generate sustained best-so-far improvements, whereas weaker models exhibit semantic drift and stagnation. However, at operator-level, this suggests that beyond base capability, effective LLM optimizers behave as \textbf{local refiners}: they frequently produce offspring that strictly improve upon their prompted parents while maintaining a controlled semantic step sizes. We then validate this hypothesis through two studies below.


\subsubsection{Model-Level Regression}

We first conduct a fine-grained regression at model level. We employ two operator-level metrics defined at the parent$\rightarrow$child mutation step. First, the \textbf{local refinement rate (LRR)} represents the frequency of strict improvements of the offspring over prompted parents at the fraction of valid offspring attempts. Second, the \textbf{parent--child distance (PCD)} quantifies the average semantic distance between each offspring and its prompted parents in the same task-specific semantic space. 



Table~\ref{tab:lrr_pcd_absorption} reveals that when considered alone, larger semantic step sizes (PCD) are negatively correlated with final performance. However, this effect vanishes once LRR is included, while LRR remains strongly positive and highly significant. This is also consistent with our interaction finding: larger edits tend to reduce the probability of producing refinements. In other words, the negative effect of large semantic modifications is largely explained by their impact on refinement reliability.

This operator-level regression once again highlights that good LLM optimizers act as \textbf{local refiners}, where performance is governed by the ability to produce reliable incremental improvements rather than by the magnitude of semantic variation.

\begin{table}[tbp]
    \centering
    \scriptsize
    \begin{tabular}{lcc}
    \toprule
     & \textbf{ZS + PCD} & \textbf{ZS + LRR + PCD} \\
    \midrule
    \textbf{Zero-shot Perf. (z) }
     & \cellcolor{myGreen!44} $0.233^{*}$ & \cellcolor{myGreen!22} $0.144$ \\
     & (0.028) & (0.112) \\[0.4ex]
    \textbf{Avg. Parent--Child Distance (z) }
     & \cellcolor{myRed!66} $-0.329^{**}$ & \cellcolor{myRed!11} $-0.024$ \\
     & (0.001) & (0.838) \\[0.4ex]
    \textbf{Avg. Local Refinement Rate (z)}
     &  & \cellcolor{myGreen!88} $0.528^{***}$ \\
     &  & ($<$0.001) \\
    \midrule
    $R^2$ & 0.204 & 0.367 \\
    \bottomrule
    \end{tabular}
\caption{Model--task OLS regressions predicting best final performance (z-score), with task fixed effects and model-clustered standard errors. p-values in parentheses. $^{***}p<0.001$, $^{**}p<0.01$,$^{*}p<0.05$. Cells are shaded by coefficient magnitude.}
    \label{tab:lrr_pcd_absorption}
\end{table}

\subsubsection{Perturbation Study: Model Mixing}
To provide interventional evidence of the role of local refinement behavior, we further perform a perturbation study through model mixing experiments. 

At each generation, a fraction of offspring are generated by an alternative model (weak refiner), while the remaining offspring are produced by the primary model (strong refiner). This intervention directly manipulate the refinement behavior of the search process. We construct task-specific model pairs with comparable zero-shot performance but contrasting refinement capabilities (see Appendix Figure~\ref{fig:refine_model}). We evaluate this intervention across three representative sub-tasks across different regimes: TSP-60, bin packing-OR3, and Prompt Optimization-Summarization.

\begin{figure}[tbp]
    \centering
    \includegraphics[width=\columnwidth]{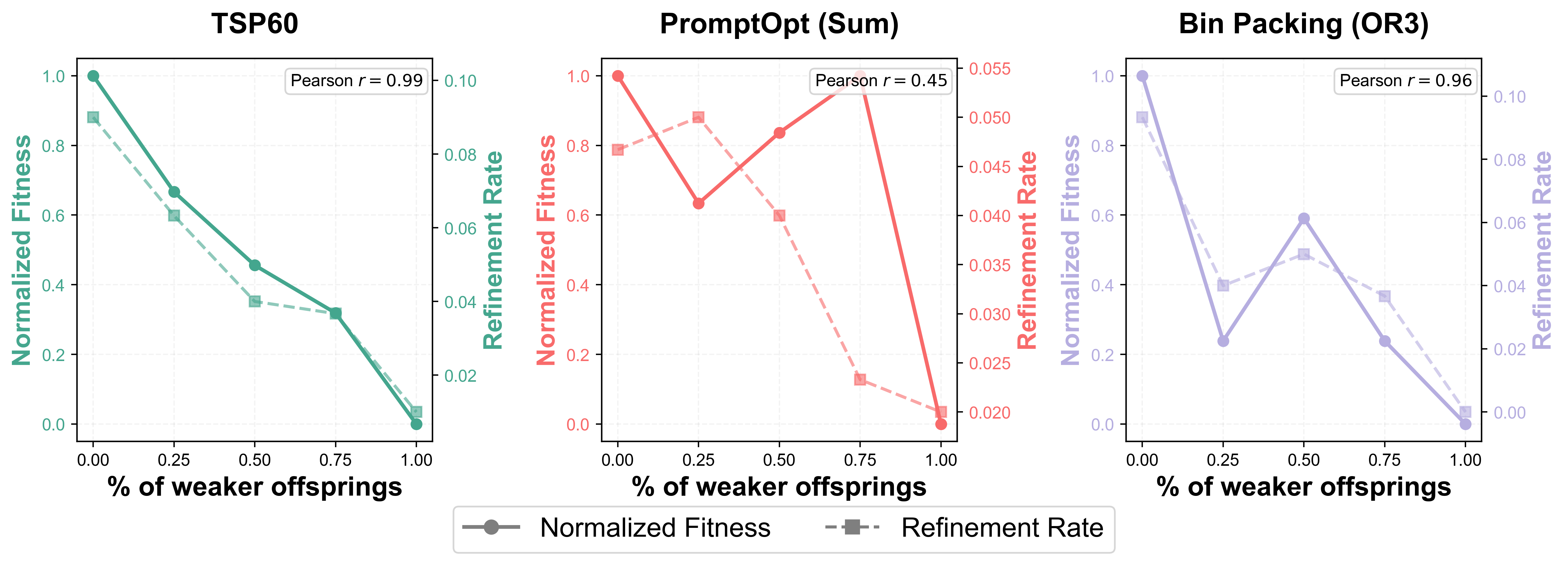}
    \caption{\textbf{Effect of model mixing on optimization performance and refinement rate.} A fraction of offspring is generated by a weaker refiner. Solid lines denote fitness; dashed lines denote refinement rate.}
    \label{fig:model_mixing}
\end{figure}

As shown in Figure~\ref{fig:model_mixing}, as the proportion of weak-refiner offspring increases, in particular, on TSP-60 and bin packing tasks, performance degrades sharply and monotonically. The same effect exists yet appears to be a bit weaker and less consistent in prompt optimization. Moreover, higher weaker offspring's ratios consistently reduce the overall refinement rate, which co-varies with the observed degradation in performance. These results suggest that weakening refinement behavior, by injecting lower-refinement operators, could impair the system's ability to produce sustained improvements, therefore leading to worse optimization outcomes.

\subsection{Cost-Efficiency Implication}\label{Sec:cost_efficiency}

\begin{figure}[htbp]  
    \centering
    \includegraphics[width=0.9\columnwidth]{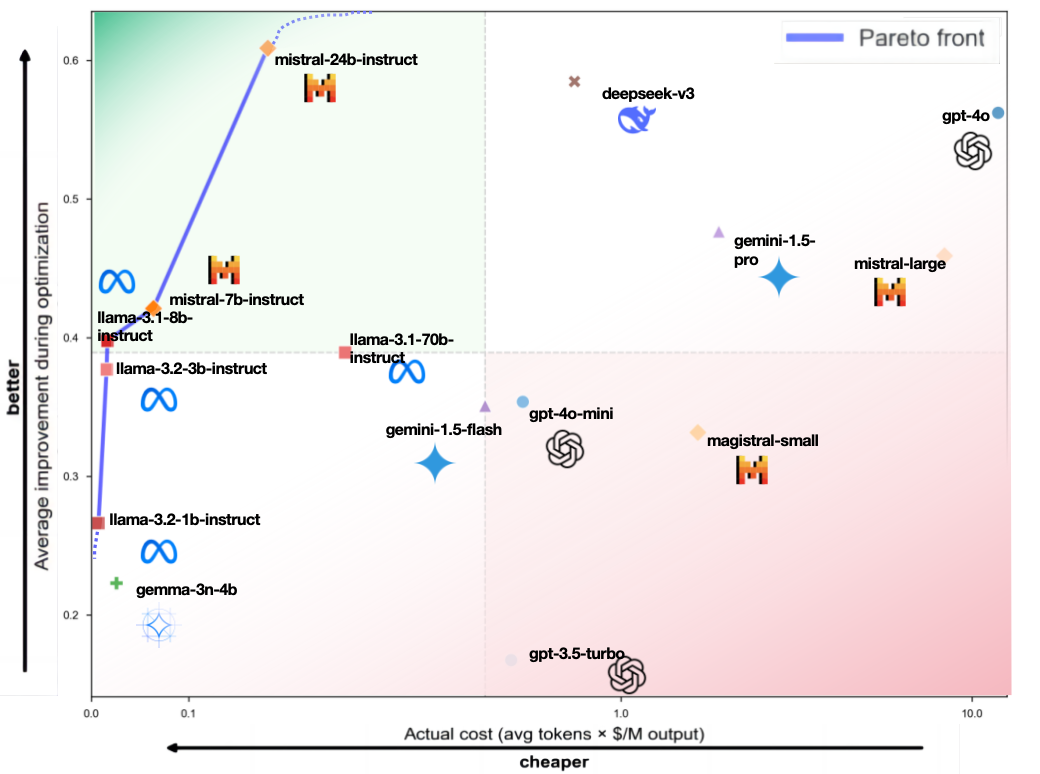}
    \caption{\textbf{Optimization gain versus cost across LLMs.} Each point represents a model, plotting average fitness improvement achieved through evolution against estimated monetary cost.}
    \label{fig:cost_analysis}
\end{figure}

Our findings also carry practical implications for cost-sensitive deployment. Since optimization performance is not fully determined by base model capability, strong optimizers are not necessarily the most expensive models. We thus estimate the monetary cost of evolutionary optimization for each model based on the average number of input and output tokens per run, using API pricing in OpenRouter platform~\footnote{\url{https://openrouter.ai}}. Optimization efficacy is measured as the fitness gain achieved over evolution. Figure~\ref{fig:cost_analysis} situates all models in a cost-improvement space, aggregated across all tasks (See Figure~\ref{fig:cost_four_task}).

This aggregated view reveals large variation in cost--performance trade-offs. Notably, some mid-sized models achieve large fitness improvements at relatively low cost, whereas stronger zero-shot models do not always yield proportional gains per dollar. For example, Mistral-24B-Instruct lies on the Pareto frontier, combining large fitness improvement with moderate cost, and thus represents an efficient optimization operator rather than merely a strong base model. Overall, this result reinforces our central claim: effective evolutionary optimization depends more on how a model refines solutions over time than on its raw problem-solving capability. For practitioners, this can help build cost-efficient evolutionary systems by selecting models with favorable optimization behavior, instead of defaulting to the most powerful LLM.




\section{Discussion and Conclusion}


In this work, we examined the exploration--exploitation trade-off in LLM-guided evolutionary search to understand why some models act as substantially better search operators than others. Although zero-shot task performance correlates with final optimization outcomes, it explains only part of the variance: models with similar zero-shot capability can induce markedly different optimization trajectories and final fitness.

Compared to classical evolutionary algorithms relying on stochastic mutation/crossover and selection to balance exploration and exploitation, LLM-guided evolution alters this paradigm. The mutation operator is no longer random but instantiated by a learned generative prior that induces structured, semantically meaningful variations, thereby strongly biasing the search toward exploitation.

%


A natural hypothesis is that this lack of randomness makes novelty or diversity a bottleneck, such that increased exploration should improve performance. Our results contradict this view. Higher novelty is not systematically associated with better outcomes and often signals failure: ineffective operators drift across semantic space without refining promising solutions. In contrast, strong LLM operators behave as effective local refiners. Their trajectories progressively localize around high-performing regions, with LLMs producing frequent, incremental improvements. In this regime, novelty is beneficial only when deviations occur within already promising regions.  Our perturbation experiments further validate that directly degrading refinement behavior through model mixing brings about drops in optimization performance. These results also offer a refined interpretation of the role of novelty in LLM-guided search. Rather than being a stochastic explorer, novelty acts as an immediate driver of exploratory breakthroughs, but more importantly its long-term utility depends on whether the search regime allows these deviations to be selectively retained and amplified.


However, the observed local refinement behavior should not be viewed as an inherent capability of the base model alone. Instead, it emerges as a property of the entire agentic system that generates offspring, including the model, the prompting strategy, and the decoding configuration. While changes in temperature affect both refinement rates and performance, the relationship between refinement behavior and performance remains stable across a range of settings (see Appendix~\ref{sec:temp_sensi}).


Our findings have direct implications for the design of LLM-based optimization systems. Rather than focusing solely on maximizing base model capability, our results suggest that the key objective is to \emph{control and optimize refinement behavior}. First, stronger base models do not necessarily yield better search operators, smaller or cheaper LLMs can outperform larger ones when their inductive biases favor stable local refinement. This highlights the importance of model selection. Besides, refinement behavior can potentially be modulated via system-level design choices, including prompting and decoding hyperparameter. More broadly, our results support training or fine-tuning LLMs as effective operators~\citep{brahmachary2024largelanguagemodelbasedevolutionary,surina2025algorithm}, emphasizing local refinement and error correction rather than general-purpose capability. Understanding and shaping such operator-specific behaviors is a promising avenue for boosting LLM-guided optimization.


Finally, the geometric analysis framework we developed to study LLM-guided optimization trajectories is broadly applicable and can be repurposed to analyze other types of iterative search or agentic behaviors. For illustration, a rich, interactive collection of visualizations showing trajectories in semantic space across models and tasks can be found on our project website.

\section*{Limitations}Our study is subject to several limitations. First, while we conduct robustness analyses on decoding hyperparameters (notably temperature; see Appendix~\ref{sec:temp_sensi}), our experiments still rely on a fixed evolutionary protocol. Other design choices, such as selection pressure, offspring size, and alternative sampling strategies, may influence the balance between exploration and exploitation, and could further shape cross-model differences. Second, in our study novelty is primarily operationalized as nearest-neighbor distance. Broader comparisons to KNN/average-distance novelty and alternative diversity indices would help assess robustness. Third, although we include perturbation experiments via model mixing, the intervention is still hard to fully isolate local refinement in a strictly controlled manner. Replacing the model that generates offspring may also affect other latent and invisible characteristics (e.g., reasoning patterns or exploration tendencies), thus 
making it difficult to attribute all performance differences solely to local refinement.


\section*{Acknowledgments}
This work was supported by ANR (grant ANR-22-CPJ2-0036-01). It was also partially supported by ANR through the MIAI "AI \& Language" chair (ANR-23-IACL-0006).


\bibliography{main}


\appendix


\section{Complete Experimental Result}
\label{sec:appendix_result}

See Table~\ref{tab:results_rq1}.

\begin{table*}[htbp]
\centering
\small
\setlength{\tabcolsep}{6.1pt}  
\begin{tabular}{
  l 
  *{5}{>{\centering\arraybackslash}p{1.05em}}  
  *{5}{>{\centering\arraybackslash}p{1.05em}}  
  *{5}{>{\centering\arraybackslash}p{1.05em}}  
}
\toprule

\multirow{2}{*}{\thead{\large{\textbf{LLM}}}}
& \multicolumn{5}{c}{\thead{\textbf{Zero-Shot}}}
& \multicolumn{5}{c}{\thead{\textbf{First Generation}}}
& \multicolumn{5}{c}{\thead{\textbf{Last Generation}}} \\
\cmidrule(lr){2-6} \cmidrule(lr){7-11} \cmidrule(lr){12-16}
& \multicolumn{1}{c}{\icon{route}} 
& \multicolumn{1}{c}{\icon{text}} 
& \multicolumn{1}{c}{\icon{math}} 
& \multicolumn{1}{c}{\icon{code}}
& \multicolumn{1}{c}{Avg}
& \multicolumn{1}{c}{\icon{route}} 
& \multicolumn{1}{c}{\icon{text}} 
& \multicolumn{1}{c}{\icon{math}} 
& \multicolumn{1}{c}{\icon{code}}
& \multicolumn{1}{c}{Avg}
& \multicolumn{1}{c}{\icon{route}} 
& \multicolumn{1}{c}{\icon{text}} 
& \multicolumn{1}{c}{\icon{math}} 
& \multicolumn{1}{c}{\icon{code}}
& \multicolumn{1}{c}{Avg}\\
\midrule

\icon{openai}~ 4o & \textbf{47.4} & 51.9 & \textbf{82.3} & 31.5 & \textbf{53.3} & \cellcolor{myGreen!9} 37.2 & \cellcolor{myGreen!10} 41.3 & \cellcolor{myGreen!41} 75.7 & \cellcolor{myGreen!8} 31.7 & \cellcolor{myGreen!13} 46.5 & \cellcolor{myGreen!100}{85.4} & \cellcolor{myGreen!52} 70.9 & \cellcolor{myGreen!64} 77.7 & \cellcolor{myGreen!90}{75.4} & \cellcolor{myGreen!77} 77.4 \\
\icon{gemini}~ 1.5-Pro & 47.3 & 43.0 & 70.4 & 30.7 & 47.8 & \cellcolor{myGreen!14} 43.0 & \cellcolor{myGreen!9} 38.3 & \cellcolor{myGreen!48} 84.0 & \cellcolor{myGreen!8} 32.4 & \cellcolor{myGreen!15} 49.4 & \cellcolor{myGreen!100}{\textbf{89.0}} & \cellcolor{myGreen!45} 72.4 & \cellcolor{myGreen!88}{85.5} & \cellcolor{myGreen!39} 58.5 & \cellcolor{myGreen!69} 76.4 \\
\icon{deepseek}~ V3 & 39.0 & 41.2 & 71.5 & 31.5 & 45.8 & \cellcolor{myGreen!24} \textbf{50.8} & \cellcolor{myGreen!22} 56.6 & \cellcolor{myGreen!60} \textbf{87.2} & \cellcolor{myGreen!8} 33.0 & \cellcolor{myGreen!25} \textbf{56.9} & \cellcolor{myGreen!64} 70.4 & \cellcolor{myGreen!81}{77.1} & \cellcolor{myGreen!100}{91.1} & \cellcolor{myGreen!51} 62.7 & \cellcolor{myGreen!75} 75.3 \\
\icon{mistral}~ Large & 19.5 & 49.1 & 79.7 & 31.5 & 45.0 & \cellcolor{myGreen!8} 34.5 & \cellcolor{myGreen!21} 56.6 & \cellcolor{myGreen!14} 74.1 & \cellcolor{myGreen!8} 33.0 & \cellcolor{myGreen!11} 49.5 & \cellcolor{myGreen!37} 58.4 & \cellcolor{myGreen!100}{84.7} & \cellcolor{myGreen!44} 78.7 & \cellcolor{myGreen!85}{81.1} & \cellcolor{myGreen!64} 75.7 \\
\icon{meta}~ 3.1-70B-Instruct & 15.9 & 55.2 & 75.2 & 31.5 & 44.5 & \cellcolor{myGreen!8} 34.0 & \cellcolor{myGreen!14} 45.1 & \cellcolor{myGreen!17} 69.9 & \cellcolor{myGreen!8} 33.0 & \cellcolor{myGreen!10} 45.5 & \cellcolor{myGreen!40} 59.6 & \cellcolor{myGreen!42} 69.5 & \cellcolor{myGreen!66} 78.0 & \cellcolor{myGreen!64} 69.8 & \cellcolor{myGreen!52} 69.2 \\
\icon{mistral}~ Magistral-Small & 29.0 & 49.9 & 66.6 & 31.5 & 44.3 & \cellcolor{myGreen!8} 34.0 & \cellcolor{myGreen!27} 61.8 & \cellcolor{myGreen!19} 70.5 & \cellcolor{myGreen!8} 31.7 & \cellcolor{myGreen!12} 49.5 & \cellcolor{myGreen!58} 68.0 & \cellcolor{myGreen!53} 73.7 & \cellcolor{myGreen!36} 75.4 & \cellcolor{myGreen!54} 64.4 & \cellcolor{myGreen!50} 70.4 \\
\icon{mistral}~ 24B-Instruct & 11.6 & \textbf{55.8} & 72.2 & 31.5 & 42.8 & \cellcolor{myGreen!8} 34.0 & \cellcolor{myGreen!32} \textbf{66.5} & \cellcolor{myGreen!17} 70.2 & \cellcolor{myGreen!8} 33.1 & \cellcolor{myGreen!13} 51.0 & \cellcolor{myGreen!76} 75.0 & \cellcolor{myGreen!75} 84.3 & \cellcolor{myGreen!42} 75.2 & \cellcolor{myGreen!100}{\textbf{92.0}} & \cellcolor{myGreen!78}{\textbf{81.6}} \\
\icon{openai}~ 4o-mini & 13.9 & 38.3 & 70.1 & 31.5 & 38.4 & \cellcolor{myGreen!8} 34.0 & \cellcolor{myGreen!21} 55.8 & \cellcolor{myGreen!8} 66.7 & \cellcolor{myGreen!8} 31.7 & \cellcolor{myGreen!9} 47.1 & \cellcolor{myGreen!41} 60.2 & \cellcolor{myGreen!40} 71.2 & \cellcolor{myGreen!62} 82.9 & \cellcolor{myGreen!51} 66.2 & \cellcolor{myGreen!48} 70.1 \\
\icon{meta}~ 3.1-8B-Instruct & 2.9 & 55.7 & 48.7 & \textbf{32.9} & 35.1 & \cellcolor{myGreen!8} 34.3 & \cellcolor{myGreen!12} 44.5 & \cellcolor{myGreen!11} 67.4 & \cellcolor{myGreen!8} 31.7 & \cellcolor{myGreen!9} 44.5 & \cellcolor{myGreen!49} 63.8 & \cellcolor{myGreen!50} 73.7 & \cellcolor{myGreen!41} 77.1 & \cellcolor{myGreen!87}{74.5} & \cellcolor{myGreen!56} 72.2 \\
\icon{mistral}~ 7B-Instruct & 18.5 & 46.9 & 49.7 & 23.6 & 34.7 & \cellcolor{myGreen!8} 35.1 & \cellcolor{myGreen!19} 47.7 & \cellcolor{myGreen!14} 67.7 & \cellcolor{myGreen!8} 31.7 & \cellcolor{myGreen!11} 45.5 & \cellcolor{myGreen!18} 46.9 & \cellcolor{myGreen!51} 73.7 & \cellcolor{myGreen!92}{\textbf{91.8}} & \cellcolor{myGreen!74} 67.7 & \cellcolor{myGreen!55} 70.0 \\
\icon{gemini}~ 1.5-Flash & 6.6 & 47.5 & 60.9 & 3.4 & 29.6 & \cellcolor{myGreen!8} 34.0 & \cellcolor{myGreen!15} 49.8 & \cellcolor{myGreen!33} 73.5 & \cellcolor{myGreen!8} 31.7 & \cellcolor{myGreen!13} 47.2 & \cellcolor{myGreen!35} 57.1 & \cellcolor{myGreen!100}{\textbf{95.9}} & \cellcolor{myGreen!36} 75.1 & \cellcolor{myGreen!15} 44.6 & \cellcolor{myGreen!48} 68.2 \\
\icon{meta}~ 3.2-3B-Instruct & 22.2 & 36.3 & 28.4 & 29.8 & 29.1 & \cellcolor{myGreen!8} 34.6 & \cellcolor{myGreen!17} 47.7 & \cellcolor{myGreen!8} 66.7 & \cellcolor{myGreen!8} 32.3 & \cellcolor{myGreen!9} 45.3 & \cellcolor{myGreen!31} 55.1 & \cellcolor{myGreen!51} 70.6 & \cellcolor{myGreen!98}{85.9} & \cellcolor{myGreen!26} 47.5 & \cellcolor{myGreen!49} 64.8 \\
\icon{meta}~ 3.2-1B-Instruct & 14.8 & 47.8 & 0.0 & 31.5 & 23.5 & \cellcolor{myGreen!8} 34.0 & \cellcolor{myGreen!18} 53.1 & \cellcolor{myGreen!17} 69.4 & \cellcolor{myGreen!8} 31.7 & \cellcolor{myGreen!11} 47.0 & \cellcolor{myGreen!29} 54.1 & \cellcolor{myGreen!43} 68.5 & \cellcolor{myGreen!67} 80.9 & \cellcolor{myGreen!25} 48.5 & \cellcolor{myGreen!40} 63.0 \\
\icon{gemma}~ 3n-4B & 0.0 & 55.6 & 0.0 & 22.9 & 19.6 & \cellcolor{myGreen!8} 35.6 & \cellcolor{myGreen!18} 53.2 & \cellcolor{myGreen!8} 66.7 & \cellcolor{myGreen!8} 31.7 & \cellcolor{myGreen!9} 46.8 & \cellcolor{myGreen!13} 42.7 & \cellcolor{myGreen!52} 80.4 & \cellcolor{myGreen!11} 67.8 & \cellcolor{myGreen!34} 52.2 & \cellcolor{myGreen!25} 60.8 \\
\icon{openai}~ 3.5-turbo & 16.5 & 27.2 & 0.0 & 28.1 & 18.0 & \cellcolor{myGreen!8} 34.0 & \cellcolor{myGreen!10} 40.5 & \cellcolor{myGreen!32} 75.6 & \cellcolor{myGreen!8} \textbf{33.2} & \cellcolor{myGreen!12} 45.8 & \cellcolor{myGreen!21} 49.0 & \cellcolor{myGreen!42} 65.8 & \cellcolor{myGreen!60} 81.4 & \cellcolor{myGreen!12} 41.0 & \cellcolor{myGreen!31} 59.3 \\

\bottomrule
\end{tabular}
\caption{Fitness performance of LLMs on four task families (\icon{route}~ Route Optimization, \icon{text}~ Prompt Optimization, \icon{math}~ Equation Discovery, \icon{code}~ Heuristic Design). Cells report averaged normalized fitness across two sub-tasks and two seeds within the same family, making scores comparable over three settings. Those cells are background-shaded by a normalized improvement (darker = larger improvement) computed per \texttt{(model,task)} pair relative to their initial-population best value of each sub-task.
Models are sorted in descending order of their average Zero-Shot score. Best scores per column are \textbf{bold}.}
\label{tab:results_rq1}
\end{table*}

\section{Use of AI Assistant}
AI tools were used to assist in writing, editing,
and code development. The authors provided all
content, ideas, and decisions, and the AI was used
solely to improve clarity, readability, and efficiency.

\section{Task-Specific Experimental Details}\label{evolution_protocol_details}
Here are task-family-specific details of (i) EA parameters, (ii) genome validity checks, (iii) fitness evaluation, (iv) novelty distance, (v) population initialization, and (vi) prompts used for zero-shot and evolutionary search. Unless stated otherwise, the global evolutionary loop (selection--mutation--evaluation--pool update) follows Section~\ref{sec:protocol}.

\subsection{\adjustbox{valign=b}{\includegraphics[height=1.3em]{img/icons/route.png}}~ Route Optimization}
\label{app:tsp}

\paragraph{EA Parameters:} For both TSP-30 and TSP-60, we use: $n_{\mathrm{init}}=40$ (initial population size), $q=0.2$ (elite fraction), $p_{\mathrm{parent}}=3$ (parents sampled per generation), $p_{\mathrm{child}}=10$ (offspring per generation), $N=40$ (capacity-limited pool), $G=30$ (generations), and seed $=21$.

\paragraph{Genome and Validity:} A genome is a permutation $\pi \in S_n$ where $n \in \{30, 60\}$ is the number of cities. The LLM generates a genome as a JSON array of integers. Invalid genomes (non-permutations, unparsable outputs) receive fitness $f=0$ and are excluded from parent sampling.

\paragraph{Fitness Evaluation:} Given a distance matrix $\mathrm{DIST} \in \mathbb{R}^{n \times n}$, the tour length is $L(\pi) = \sum_{i=1}^{n-1} \mathrm{DIST}_{\pi_i, \pi_{i+1}} + \mathrm{DIST}_{\pi_n, \pi_1}$. Fitness is the inverted length: $f_{\mathrm{TSP}}(\pi) = -L(\pi)$ (normalized post-hoc by task-level min/max).

\paragraph{Novelty Distance:} We use edge-set distance after canonization:
$$D_{\mathrm{edge}}(\pi, \sigma) = 1 - \frac{|E(\pi) \cap E(\sigma)|}{|E(\pi)|},$$
where $E(\pi)$ denotes the set of undirected edges in tour $\pi$. This metric captures structural differences regardless of rotation or starting city.

\paragraph{Population Initialization:} The initial population $\mathcal{P}_0$ consists of $n_{\mathrm{init}}=40$ random permutations (using the same random seed across all models). Each permutation is obtained via \texttt{random.sample(range(n), n)}.

\paragraph{Zero-shot Prompt:} The zero-shot prompt provides the complete distance matrix as JSON and asks the model to return an optimal tour. See Table~\ref{tab:prompt-tsp-zeroshot}.

\begin{table*}[t]
\centering
\begin{tcolorbox}[
  colback=myGreen!5,
  colframe=myGreen!75,
  title=TSP Zero-shot Prompt,
  fonttitle=\bfseries,
  ]
\small
\noindent
\textbf{System:} You are an optimization expert helping to solve a hard problem. You will be shown several candidate solutions with their scores. Your goal is to propose better solutions. \\[0.5ex]
\textbf{User:} 
TASK DESC: The traveling salesman problem (TSP) aims to find the shortest route visiting all cities exactly once. You must return a valid tour as a list of city indices.\\[0.3ex]
QUESTION: [Distance matrix provided as JSON]\\[0.3ex]
Please return the optimal solution as JSON without extra explanation: \texttt{\{ "genome": "[list of n unique integers 0 to n-1]" \}}.

\end{tcolorbox}
\caption{TSP Zero-shot Prompt Template}
\label{tab:prompt-tsp-zeroshot}
\end{table*}

\paragraph{Evolution Prompt:} During evolution, the prompt provides the distance matrix and up to 3 parent tours with their scores (path lengths). The LLM is asked to generate a better child tour. See Table~\ref{tab:prompt-tsp-evolve}.

\begin{table*}[t]
\centering
\begin{tcolorbox}[
  colback=myGreen!5,
  colframe=myGreen!75,
  title=TSP Evolution Prompt,
  fonttitle=\bfseries,
  ]
\small
\noindent
\textbf{System:} You are an optimization expert helping to solve a hard problem. You will be shown several candidate solutions with their scores. Your goal is to propose better solutions. \\[0.5ex]
\textbf{User:} 
TASK DESC: The traveling salesman problem (TSP) aims to find the shortest route visiting all cities exactly once.\\[0.3ex]
QUESTION: [Distance matrix provided as JSON]\\[0.3ex]
Here are [num\_parents] previous solutions and their scores (lower is better):
\begin{verbatim}
{"genome": [parent_1], "score": 1234.5}
{"genome": [parent_2], "score": 1256.3}
...
\end{verbatim}
Please return one BETTER child genome as JSON: \texttt{\{ "genome": "[full-new]" \}}. The genome must be a list of [n] unique integers from 0 to [n-1].

\end{tcolorbox}
\caption{TSP Evolution Prompt Template}
\label{tab:prompt-tsp-evolve}
\end{table*}

\subsection{\adjustbox{valign=b}{\includegraphics[height=1.3em]{img/icons/text.png}}~ Prompt Optimization}
\label{app:promptopt}

\paragraph{EA Parameters:}  $n_{\mathrm{init}}=10$, $q=0.2$, $p_{\mathrm{parent}}=2$, $p_{\mathrm{child}}=5$, $N=10$ (capacity), $G=30$, and two task variants: \textbf{SAMSum} (dialogue summarization) and \textbf{ASSET} (text simplification).

\paragraph{Genome and Validity:} A genome is a natural language instruction prompt (string). Any non-empty string output is considered valid.

\paragraph{Fitness Evaluation:} For each prompt genome $p$, we condition a frozen evaluator LLM (gpt-4o-mini) on $p$ to generate outputs on a held-out 25\% evaluation set. Fitness is the task-specific metric score (ROUGE-L for SAMSum, SARI for ASSET).

\paragraph{Novelty Distance:} We use semantic cosine distance in a fixed embedding space (OpenAI's \texttt{text-embedding-ada-002}):
$$D_{\mathrm{cos}}(p_1, p_2) = 1 - \frac{E(p_1) \cdot E(p_2)}{\|E(p_1)\| \|E(p_2)\|},$$
where $E(\cdot)$ is the sentence embedding.

\paragraph{Population Initialization:} Initial prompts are loaded from a curated set of baseline prompts for each task (SAMSum or ASSET). Duplicates are removed, and the first 10 unique prompts are selected.

\paragraph{Zero-shot Prompt:} See Table~\ref{tab:prompt-promptopt-zeroshot}.

\begin{table*}[t]
\centering
\begin{tcolorbox}[
  colback=myGreen!5,
  colframe=myGreen!75,
  title=Prompt Optimization Zero-shot Prompt,
  fonttitle=\bfseries,
  ]
\small
\noindent
\textbf{System:} You are a prompt optimization expert. Your goal is to design effective prompts for language models on text summarization/simplification tasks. \\[0.5ex]
\textbf{User:} 
TASK DESC: Your task is to create a high-quality instruction prompt that will guide a language model to perform [summarization/simplification] with high fidelity to the original content.\\[0.3ex]
Please return ONLY the prompt text (plain text, not JSON or wrapped in quotes) that will be used to instruct the language model.\\[0.3ex]
Example expected output:
\begin{verbatim}
Summarize the following dialogue in a concise manner...
\end{verbatim}

\end{tcolorbox}
\caption{Prompt Optimization Zero-shot Prompt Template}
\label{tab:prompt-promptopt-zeroshot}
\end{table*}

\paragraph{Evolution Prompt:} See Table~\ref{tab:prompt-promptopt-evolve}.

\begin{table*}[t]
\centering
\begin{tcolorbox}[
  colback=myGreen!5,
  colframe=myGreen!75,
  title=Prompt Optimization Evolution Prompt,
  fonttitle=\bfseries,
  ]
\small
\noindent
\textbf{System:} You are a prompt optimization expert. Your goal is to design effective prompts for language models. \\[0.5ex]
\textbf{User:} 
TASK DESC: Create a better instruction prompt for [summarization/simplification].\\[0.3ex]
Here are previous prompts and their performance scores (higher fitness = better):
\begin{verbatim}
Prompt: "Summarize the following..."
Fitness: 0.1234
Prompt: "Create a brief summary..."
Fitness: 0.1156
\end{verbatim}
Analyze the patterns in successful prompts. Combine successful elements and create a new, improved prompt. Return ONLY the plain text prompt without explanation.

\end{tcolorbox}
\caption{Prompt Optimization Evolution Prompt Template}
\label{tab:prompt-promptopt-evolve}
\end{table*}

\subsection{\adjustbox{valign=b}{\includegraphics[height=1.3em]{img/icons/math.png}}~ Equation Discovery}
\label{app:symboreg}

\paragraph{EA Parameters:} $n_{\mathrm{init}}=7$, $q=0.2$, $p_{\mathrm{parent}}=2$, $p_{\mathrm{child}}=10$, $N=40$ (capacity), $G=30$, and seed $=21$. We evaluate on two nonlinear oscillator datasets with different dimensionalities (Oscillator-1: 3 variables; Oscillator-2: 4 variables with time).

\paragraph{Genome and Validity:} A genome is a Python function string defining $a = f(x, v)$ (Oscillator-1) or $a = f(t, x, v)$ (Oscillator-2). Genomes are validated by attempting to parse and execute them; non-executable or divergent outputs receive fitness $f = 1 \times 10^6$ (high loss).

\paragraph{Fitness Evaluation:} Given training data $(X, y_{\mathrm{true}})$, fitness is computed as:
$$f_{\mathrm{SymReg}} = 1 - \mathrm{norm}\left(\mathrm{MSE}\left(y_{\mathrm{pred}}, y_{\mathrm{true}}\right)\right),$$
where normalization is per-task instance.

\paragraph{Novelty Distance:} We use functional behavior distance over a fixed input grid:
$$D_{\mathrm{sem}}(f, g) = 1 - \frac{1}{m} \sum_{j=1}^m \cos\left(f(x_j), g(x_j)\right),$$
where $x_j$ are uniformly sampled input points and $\cos(\cdot)$ is cosine similarity. This captures semantic divergence in output behavior.

\paragraph{Population Initialization:} Initial genomes are randomly sampled expressions combining input variables, constants, and allowed functions (e.g., \texttt{np.sin}, \texttt{np.cos}, \texttt{np.exp}). We maintain the same seed for all models. All initial genomes are evaluated on the training set.

\paragraph{Zero-shot Prompt:} See Table~\ref{tab:prompt-symreg-zeroshot}.

\begin{table*}[t]
\centering
\begin{tcolorbox}[
  colback=myGreen!5,
  colframe=myGreen!75,
  title=Symbolic Regression Zero-shot Prompt,
  fonttitle=\bfseries,
  ]
\small
\noindent
\textbf{System:} You are a scientific equation discovery expert. Your goal is to propose symbolic expressions that fit the data well. \\[0.5ex]
\textbf{User:} 
TASK DESC: This is a symbolic regression task for a damped nonlinear oscillator. Given (x, v) and acceleration a, find a mathematical expression $a = f(x, v)$ that fits the data as accurately as possible.\\[0.3ex]
QUESTION: Here are some data points [100 samples as JSON]:
\begin{verbatim}
[{"x": 0.123, "v": 0.456, "a": 0.789}, ...]
\end{verbatim}
Please return a Python function as a code block without explanation:
\begin{verbatim}
def equation(x, v):
    return 1.2*x + 0.8*v + np.sin(x)
\end{verbatim}

\end{tcolorbox}
\caption{Symbolic Regression Zero-shot Prompt Template}
\label{tab:prompt-symreg-zeroshot}
\end{table*}

\paragraph{Evolution Prompt:} See Table~\ref{tab:prompt-symreg-evolve}.

\begin{table*}[t]
\centering
\begin{tcolorbox}[
  colback=myGreen!5,
  colframe=myGreen!75,
  title=Symbolic Regression Evolution Prompt,
  fonttitle=\bfseries,
  ]
\small
\noindent
\textbf{System:} You are a scientific equation discovery expert. Your goal is to propose a new, better mathematical expression that fits the data (lower MSE is better). \\[0.5ex]
\textbf{User:} 
TASK DESC: Symbolic regression for damped nonlinear oscillator. Find $a = f(x, v)$.\\[0.3ex]
Here are previous candidate expressions and their MSE scores:
\begin{verbatim}
{"code": "def equation(x, v): return x + v", "mse_score": 0.1234}
{"code": "def equation(x, v): return np.sin(x) + v**2", "mse_score": 0.0987}
\end{verbatim}
Please return a new, better Python function as a code block without explanation. The function should take x, v as input and may use mathematical operations and numpy functions.

\end{tcolorbox}
\caption{Symbolic Regression Evolution Prompt Template}
\label{tab:prompt-symreg-evolve}
\end{table*}

\subsection{\adjustbox{valign=b}{\includegraphics[height=1.3em]{img/icons/code.png}}~ Heuristic Design}
\label{app:binpack}

\paragraph{EA Parameters:} For bin packing heuristic design, we use: $n_{\mathrm{init}}=7$, $q=0.2$, $p_{\mathrm{parent}}=2$, $p_{\mathrm{child}}=10$, $N=40$ (capacity), $G=30$, and seed $=42$. 

\paragraph{Genome and Validity:} A genome is a Python function string implementing a priority heuristic \texttt{def priority(item, bins): ...}. The function takes item size and bin residual capacities as input, and returns priority scores. Invalid or non-executable functions receive fitness $f = 1 \times 10^6$.

\paragraph{Fitness Evaluation:} For each genome, we run online bin packing on \textbf{all} instances of the active dataset (OR3 or Weibull) using the heuristic function. Fitness is the inverted number of bins used, averaged over all instances:
$$f_{\mathrm{BinPack}} = -\frac{1}{|\mathrm{instances}|} \sum_{i} \mathrm{bins\_used}(i).$$

\paragraph{Novelty Distance:} We measure behavioral/strategy distance as well. We define a set of fixed ``probe scenarios'' (random combinations of item sizes and bin capacities), compute the priority/score vectors for each candidate function under these scenarios, and measure distance via:
$$D_{\mathrm{behav}}(h_1, h_2) = 1 - \frac{1}{K} \sum_{k=1}^{K} \cos\left(\mathbf{s}^{(k)}_{h_1}, \mathbf{s}^{(k)}_{h_2}\right),$$
where $\mathbf{s}^{(k)}_{h}$ is the priority vector returned by function $h$ on scenario $k$. This metric captures functional divergence in heuristic strategy independent of implementation style. If a function returns probability distributions, distribution distance (Jensen-Shannon or Earth Mover Distance) can alternatively be used.

\paragraph{Population Initialization:} Initial heuristics are sampled from a set of canonical bin packing rules (e.g., best-fit, worst-fit, first-fit, combinations thereof). More details can be found in our code repository. Each is evaluated on all instances.

\paragraph{Zero-shot Prompt:} See Table~\ref{tab:prompt-binpack-zeroshot}.

\begin{table*}[t]
\centering
\begin{tcolorbox}[
  colback=myGreen!5,
  colframe=myGreen!75,
  title=Bin Packing Zero-shot Prompt,
  fonttitle=\bfseries,
  ]
\small
\noindent
\textbf{System:} You are an expert in online bin packing algorithms. Your goal is to design priority functions that minimize the number of bins needed. \\[0.5ex]
\textbf{User:} 
TASK DESC: Online bin packing: given items of varying sizes and bins of fixed capacity, design a priority heuristic that decides where to place each incoming item.\\[0.3ex]
Your priority function will be called as:
\begin{verbatim}
priority(item, bins)
\end{verbatim}
Where:
\begin{itemize}
  \item \texttt{item}: float, the size of current item to pack
  \item \texttt{bins}: numpy array, remaining capacities of bins that can fit the item
\end{itemize}
Return: numpy array of same length as \texttt{bins}, with priority scores for each bin.\\[0.3ex]
Example:
\begin{verbatim}
def priority(item, bins):
    return -(bins - item)  # Best-fit heuristic
\end{verbatim}
Please provide a function better than the example! Return ONLY the code block.

\end{tcolorbox}
\caption{Bin Packing Zero-shot Prompt Template}
\label{tab:prompt-binpack-zeroshot}
\end{table*}

\paragraph{Evolution Prompt:} See Table~\ref{tab:prompt-binpack-evolve}.

\begin{table*}[t]
\centering
\begin{tcolorbox}[
  colback=myGreen!5,
  colframe=myGreen!75,
  title=Bin Packing Evolution Prompt,
  fonttitle=\bfseries,
  ]
\small
\noindent
\textbf{System:} You are an expert in online bin packing algorithms. Your goal is to design a new, better priority function that minimizes bins used. \\[0.5ex]
\textbf{User:} 
TASK DESC: Online bin packing heuristic design. Minimize the number of bins used across \textbf{all [dataset] instances}.\\[0.3ex]
Here are previous priority functions and their performance (average bins used):
\begin{verbatim}
{"code": "def priority(item, bins): return -(bins - item)", "avg_bins": 45.3}
{"code": "def priority(item, bins): return bins**2 / (item + 1e-6)", "avg_bins": 42.1}
\end{verbatim}
Analyze successful patterns and create a new, better heuristic. The function signature must be:
\begin{verbatim}
def priority(item, bins):
    import numpy as np
    # Your heuristic here
    return priority_scores
\end{verbatim}
Return ONLY the code block without explanation. Try to achieve fewer bins than the parents above!

\end{tcolorbox}
\caption{Bin Packing Evolution Prompt Template}
\label{tab:prompt-binpack-evolve}
\end{table*}

\subsubsection{Zero-shot Evaluation Details}
For each model–task pair, we sampled outputs under six temperature settings ($T\in\{0.0, 0.2, 0.4, 0.6, 0.8, 1.0\}$), with two runs per temperature. 

\subsubsection{Task-agnostic Novelty Computation}\label{app:novelty}

Algorithm~\ref{alg:novelty} details the task-agnostic novelty computation procedure used across all experiments. Novelty is defined as the minimum semantic distance to prior candidates within the same problem instance and normalized to ensure comparability across generations.

\begin{algorithm}[t]
\caption{Task-agnostic novelty computation for each genome}
\label{alg:novelty}
\SetAlgoLined
\DontPrintSemicolon
\LinesNotNumbered

\textbf{Require:} task $T$, problem instance $p$, generation $g$, candidate $a_{p,g}$, prior set $\mathcal{A}_{p,g}^\mathrm{prior}$, task-specific distance metric $D_T$, all diversity values $\mathcal{N}_p$ for instance $p$, constant $\varepsilon$ 

\textbf{Distance Computation: } Compute raw point diversity (nearest neighbor):  
\quad $n \leftarrow \min_{b \in \mathcal{A}_{p,g}^\mathrm{prior}} D_T(a_{p,g}, b)$\;

\textbf{Normalization:} Normalize within $p$:  
\quad $\widehat{n}(a_{p,g}) \leftarrow \frac{n - \min(\mathcal{N}_p)}{\max(\mathcal{N}_p) - \min(\mathcal{N}_p) + \varepsilon}$ \;

\textbf{Return} normalized diversity $\widehat{n}(a_{p,g}) \in [0,1]$
\end{algorithm}

\subsection{Multi-dimensional Scaling Parameters}\label{app:mds_params_detail}

We use Multi-dimensional Scaling (MDS) to project task-specific novelty distances into 2D embeddings (``trajectory landscapes'') for visualization. The approach is standardized across all four task families, with variations in the distance metric:

\paragraph{Sampling and Fitting Strategy:}

For each task, we collect all genomes across generations and models. To avoid computational bottlenecks:
\begin{enumerate}
  \item If total genomes $N \leq 4000$: fit MDS on the entire distance matrix.
  \item If $N > 4000$: use stratified sampling (max 60 genomes per (model, generation) bucket) to obtain $m \leq 4000$ base points, then use out-of-sample (OOS) placement for remaining points.
\end{enumerate}

\paragraph{MDS Solver Parameters:}

All experiments use \texttt{sklearn.manifold.MDS} with:
\begin{itemize}
  \item \texttt{n\_components=2}: Project to 2D for visualization.
  \item \texttt{dissimilarity="precomputed"}: Input is precomputed distance matrix.
  \item \texttt{n\_init=1}: Single initialization (fixed random seed ensures reproducibility).
  \item \texttt{max\_iter=300}: Maximum solver iterations.
  \item \texttt{eps=1e-3}: Convergence tolerance.
  \item \texttt{random\_state=42}: Deterministic seed.
\end{itemize}

\paragraph{Out-of-Sample Placement:}

For genomes not in the base fit set, we use k-NN Shepard interpolation in the 2D space:
\begin{enumerate}
  \item Compute distance (using task-specific metric) from each OOS point to all $m$ base points.
  \item Find $k=8$ nearest neighbors (smallest distances).
  \item Assign weights $w_i = 1 / (d_i + 10^{-8})^p$, where $p=2.0$ and $d_i$ is distance to neighbor $i$.
  \item Place OOS point at weighted average of neighbor 2D coordinates.
\end{enumerate}

This approach is fast (vectorized per block of 4000 points) and preserves neighborhood structure in the high-dimensional space.

\paragraph{Prompt Optimization Preprocessing:}

For prompt optimization, embeddings are high-dimensional ($\sim 1536$ dimensions for text-embedding-ada-002). To accelerate MDS on large datasets ($>10K$ rows), we first applied PyGlimmerMDS\footnote{\url{https://github.com/hageldave/PyGlimmerMDS}} (a GPU-accelerated multilevel MDS variant) on a server, then loaded precomputed 2D coordinates for visualization.

\paragraph{Normalization and Scaling:}

Fitness values are normalized per task using robust min-max scaling (1st and 99th percentiles), ensuring visual comparability across tasks with different fitness ranges. Generation and fitness are displayed via (i) color (viridis colormap) and (ii) point size, respectively.

\section{Robustness Analyses}
\label{sec:appendix_robust}

\subsection{Temperature-Sensitivity Experiment}
\label{sec:temp_sensi}
A potential concern is that our findings may depend on specific decoding hyperparameters, as temperature directly affects the stochasticity of LLM-generated mutations. We also conduct a temperature-sensitivity analysis on two representative task families—route optimization (TSP) and equation discovery (Oscillator)—using two models with contrasting refinement capabilities (Mistral-7B and Mistral-24B). We vary the decoding temperature over a wide range $T \in \{0.0, 0.1, 0.3, 0.5, 0.7, 0.9, 1.1, 1.3\}$, and for each configuration measure both the local refinement rate and the final optimization performance.

\begin{figure}[htbp]
    \centering
    \includegraphics[width=\columnwidth]{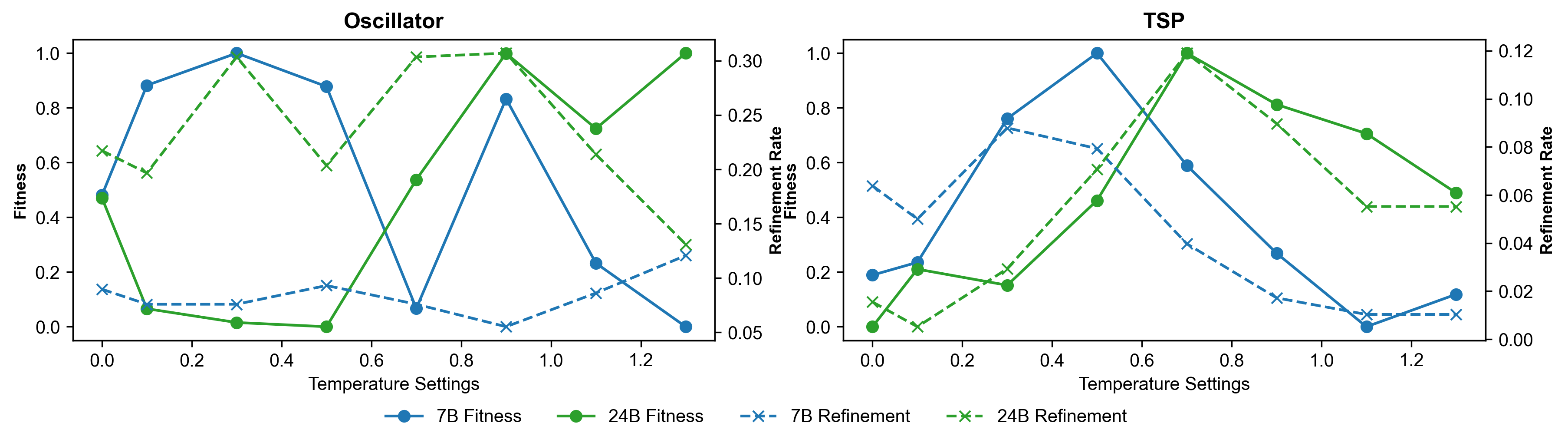}
    \caption{
Temperature sensitivity of refinement--performance dynamics. While performance varies with temperature, the relationship between refinement and fitness remains stable, particularly on TSP where strong positive correlations are observed.
}
    \label{fig:tem_exp}
\end{figure}

Overall, these additional results in Table~\ref{tab:temp_sensitivity} and Figure~\ref{fig:tem_exp} that our main finding—that local refinement ability is a key driver of optimization success—is robust to substantial variations in decoding temperature. Rather than being tied to a narrow hyperparameter regime, refinement behavior turns out to be a stable property of the combined system (model, prompt, and decoding configuration).

\begin{table}[htbp]
\centering
\footnotesize
\begin{tabular}{lccc}
\toprule
\textbf{Task} & \textbf{Model} & \textbf{Pearson $r$} & \textbf{$p$-value} \\
\midrule
Oscillator & Mistral-7B & $0.49$ & $0.209$ \\
Oscillator & Mistral-24B & $0.32$ & $0.433$ \\
\midrule
TSP & Mistral-7B & $0.76^{*}$ & $0.027$ \\
TSP & Mistral-24B & $0.92^{***}$ & $0.000$ \\
\bottomrule
\end{tabular}
\caption{
Pearson correlation between local refinement rate and final performance under temperature variation. The relationship remains consistently positive across settings.}
\label{tab:temp_sensitivity}
\end{table}

\subsection{Perturbation Study Details}
\label{sec:perturb_study_details}

\begin{figure}[!htbp]
    \centering
    \includegraphics[width=\columnwidth]{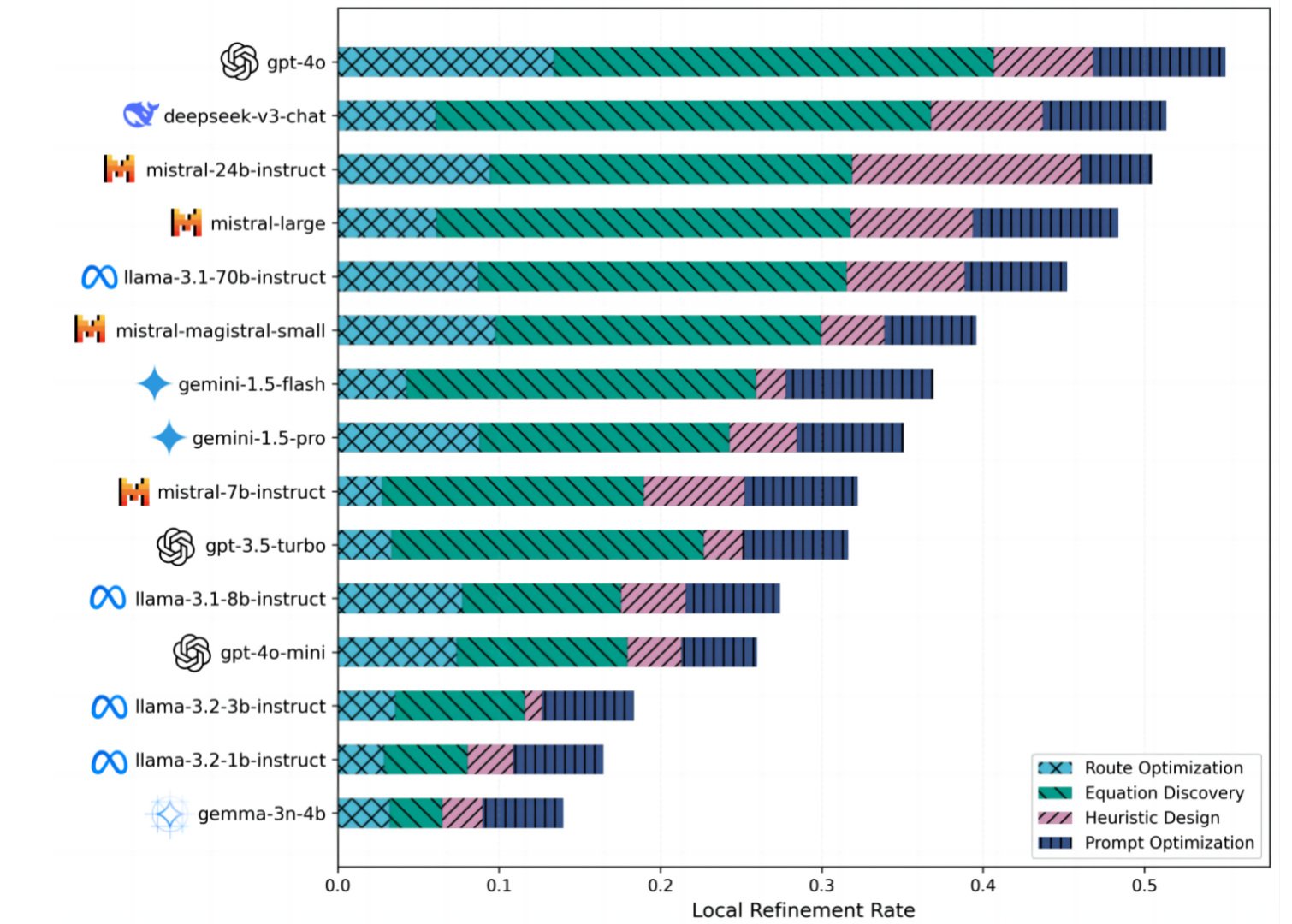}
    \caption{
Average local refinement rate across models and task families. Mistral-24B exhibits consistently strong and balanced refinement behavior across tasks, aligning with its strong post-optimization performance..}
    \label{fig:refine_model}
\end{figure}

\begin{table}[!htbp]
\centering
\scriptsize
\setlength{\tabcolsep}{3pt}  
\rowcolors{2}{gray!15}{white}
\begin{tabular}{l p{2.6cm} p{2.6cm}} 
\toprule
\textbf{Task} & \textbf{Good Refiner} & \textbf{Bad Refiner} \\
\midrule
TSP60 & Mistral-24B & Mistral-7B \\
Summarization & DeepSeek-V3-Chat & GPT-4o-mini \\
Oscillator-1 & GPT-3.5-Turbo & Gemma-3n-4B \\
Bin-Packing-OR3 & Mistral-24B & LLaMA-3.2-3B-Instruct \\
\bottomrule
\end{tabular}
\caption{Models used in the mixing model experiments across tasks. We contrast strong refiners with weaker ones to analyze causal effects of refinement ability.}
\label{tab:perturb_models}
\end{table}

\section{Statistical Model Specifications}
\label{sec:appendix_stats}

\subsection{OLS Regressions}
\label{app:ols_details}

We employ ordinary least squares (OLS) regression with clustering-robust standard errors to evaluate the predictive power of novelty and breakthrough-rate metrics on final evolutionary performance. All models include task fixed effects to account for within-task heterogeneity (8 tasks × 2 seeds = 16 task instances across 15 LLMs, $N=119$ model-task pairs).

\paragraph{Data and Estimation:}

\begin{itemize}
  \item \textbf{Sample:} Aggregated at the model-task level ($N=119$ observations: 15 models × 8 tasks, with missing cells for some model-task combinations).
  \item \textbf{Dependent variable:} $\text{best\_final\_perf}_z$: z-score-normalized best final generation fitness per (model, task) pair.
  \item \textbf{Covariates (all z-scored for interpretability)}:
    \begin{itemize}
      \item $\text{avg\_novelty}_z$: Mean within-generation novelty (average distance to nearest prior candidate).
      \item $\text{initial\_nov}_z$: Initial population novelty (diversity at generation 0).
      \item $\text{avg\_breakthrough\_rate}_z$: Fraction of generations achieving best-so-far improvement.
      \item $\text{zero\_shot\_perf}_z$: Average zero-shot performance under temperature-swept setting.
    \end{itemize}
  \item \textbf{Errors:} Clustered by model (15 clusters) using Huber-White robust covariance estimator to account for within-model correlations across tasks.
  \item \textbf{Fixed effects:} 8 task indicators (baseline: TSP-30; reference category absorbed in intercept).
\end{itemize}

\paragraph{Model Specifications:}

We fit two sets of models to test distinct hypotheses:

\textbf{Set A: Novelty as Predictor.} Regression form: $\text{best\_final\_perf}_z \sim \text{predictor}_z + \mathbf{1}_{\text{task}}$
\begin{enumerate}
  \item[M1] Predictor = $\text{avg\_novelty}_z$ only; tests if exploration (novelty) predicts final performance.
  \item[M2] Predictor = $\text{initial\_nov}_z$ only; tests if initial diversity is predictive.
  \item[M3] Predictor = $\text{zero\_shot\_perf}_z$ only; baseline model controlling for base capability.
  \item[M4] Predictors = $\text{zero\_shot\_perf}_z + \text{avg\_novelty}_z$; tests whether novelty adds explanatory power controlling for zero-shot ability.
  \item[M5] Predictors = $\text{zero\_shot\_perf}_z + \text{initial\_nov}_z$; alternative initial-diversity control.
\end{enumerate}

\textbf{Set B: Breakthrough-Rate as Predictor.} Regression form: $\text{best\_final\_perf}_z \sim \text{predictor}_z + \mathbf{1}_{\text{task}}$
\begin{enumerate}
  \item[M6] Predictor = $\text{avg\_breakthrough\_rate}_z$ only; tests if progress (breakthrough frequency) predicts success.
  \item[M7] Predictor = $\text{zero\_shot\_perf}_z$ only; baseline.
  \item[M8] Predictors = $\text{zero\_shot\_perf}_z + \text{avg\_breakthrough\_rate}_z$; joint model.
\end{enumerate}

\paragraph{Results Summary:}

See Table~\ref{tab:ols_models_a} and Table~\ref{tab:ols_models_b}

\begin{table*}[t]
\centering
\small
\caption{OLS Regression Results: Novelty Predictiveness (Set A)}
\label{tab:ols_models_a}
\begin{tabular}{lcccccc}
\toprule
\textbf{Model} & \textbf{Predictor(s)} & \textbf{$\beta$} & \textbf{SE} & \textbf{$p$-value} & \textbf{$R^2$} & \textbf{Adj.~$R^2$} \\
\midrule
M1 & avg\_novelty\_z & $-0.027$ & $0.073$ & 0.710 & 0.001 & $-0.072$ \\
M2 & initial\_nov\_z & $-0.042$ & $0.102$ & 0.681 & 0.002 & $-0.071$ \\
M3 & zero\_shot\_perf\_z & $0.322^{*}$ & $0.134$ & 0.016 & 0.103 & 0.038 \\
M4 & ZS + Avg Novelty & ZS: $0.322^{*}$ & $0.138$ & 0.019 & 0.103 & 0.029 \\
& & Nov: $0.002$ & $0.072$ & 0.980 & & \\
M5 & ZS + Init Novelty & ZS: $0.324^{*}$ & $0.132$ & 0.014 & 0.106 & 0.033 \\
& & Init: $-0.055$ & $0.063$ & 0.387 & & \\
\bottomrule
\end{tabular}
\label{novelty_prediction}
\end{table*}

\begin{table*}[t]
\centering
\small
\caption{OLS Regression Results: Breakthrough-Rate Predictiveness (Set B)}
\label{tab:ols_models_b}
\begin{tabular}{lcccccc}
\toprule
\textbf{Model} & \textbf{Predictor(s)} & \textbf{$\beta$} & \textbf{SE} & \textbf{$p$-value} & \textbf{$R^2$} & \textbf{Adj.~$R^2$} \\
\midrule
M6 & avg\_breakthrough\_rate\_z & $0.445^{***}$ & $0.097$ & $<0.001$ & 0.198 & 0.139 \\
M7 & zero\_shot\_perf\_z & $0.322^{*}$ & $0.134$ & 0.016 & 0.103 & 0.038 \\
M8 & ZS + Breakthrough & ZS: $0.226$ & $0.127$ & 0.076 & 0.246 & 0.184 \\
& & BR: $0.389^{***}$ & $0.087$ & $<0.001$ & & \\
\bottomrule
\end{tabular}
\caption*{\small $^{*}p<0.05$, $^{**}p<0.01$, $^{***}p<0.001$. All predictors are z-scored. Standard errors are clustered by model (robust to heteroskedasticity and within-model correlation). Task fixed effects included but not shown.}
\end{table*}

  
  
  
  

\paragraph{Implications:}

The weak explanatory power of novelty metrics (Set A) contrasts sharply with the strong predictiveness of breakthrough-rate metrics (Set B), supporting that models that generate diverse genomes do not necessarily improve faster. Instead, the ability to drive consistent fitness improvements (characterized by breakthrough frequency) is the key differentiator across LLMs.

\subsection{Mixed-Effects Regression Models}
\label{app:mixedlm_details}

We employ generalized linear mixed-effects models (GLMM) to analyze breakthrough probability at the generation level, accounting for model-level heterogeneity via random intercepts. We fit two specifications: (1) \textbf{Concurrent Model}: same-generation predictors, and (2) \textbf{Lagged Model}: current-generation predictors forecasting next-generation breakthrough probability.

\paragraph{Data and Estimation:}

\begin{itemize}
  \item \textbf{Concurrent sample:} $N=3{,}570$ generation-level observations across 15 LLMs (groups), with variable group sizes (min 209, max 242, mean 238). Data from all 8 tasks × 30 generations × 15 models, subset to generations with complete novelty and entropy measurements.
  \item \textbf{Lagged sample:} $N=3{,}451$ observations (omitting final generation of each model-task pair, which has no $t+1$ outcome).
  \item \textbf{Dependent variable (concurrent):} $\text{prob\_breakthrough}_z$: z-scored fraction of offspring achieving best-so-far improvement in generation $g$ (binary: 0/1 per generation, then aggregated).
  \item \textbf{Dependent variable (lagged):} $\text{prob\_breakthrough}_{z,\,t+1}$: next-generation breakthrough probability (lead variable).
  \item \textbf{Fixed effects (all z-scored for comparability)}:
    \begin{itemize}
      \item $H_{\text{fitness},\,z}$: Fitness-weighted spatial entropy (concentration of high-fitness mass).
      \item $H_{\text{spatial},\,z}$: Uniform-weighted spatial entropy (semantic dispersion).
      \item $\text{mean\_novelty\_per\_gen}_z$: Average within-generation novelty.
      \item $\text{max\_novelty\_per\_gen}_z$: Maximum novelty in generation.
      \item $\text{mean\_novelty\_per\_gen}_z \times H_{\text{spatial},\,z}$: Interaction term capturing interference between exploration and dispersion.
      \item $\text{generation}_z$: z-scored generation index (time control).
      \item 8 task indicators (baseline: TSP-30; reference absorbed in intercept).
    \end{itemize}
  \item \textbf{Random effects:} Model-level random intercept, $u_{\text{model}} \sim \mathcal{N}(0, \tau^2)$, allowing breakthrough propensity to vary across LLMs.
  \item \textbf{Estimation:} Maximum likelihood (ML, not REML) to permit likelihood ratio testing between models.
\end{itemize}

\paragraph{Model Formulation:}

\textbf{Concurrent Model.} For generation $g$ of model $m$ on task $t$:
\begin{align}
\text{prob\_breakthrough}_{g,m,t,z} = &\beta_0 + \nonumber \\
\beta_1 H_{\text{fitness},z} + \beta_2 H_{\text{spatial},z} + \nonumber \\
\beta_3 \overline{\text{nov}}_z + \beta_4 \text{max\_nov}_z + \nonumber \\
\beta_5 (\overline{\text{nov}}_z \times H_{\text{spatial},z}) \nonumber \\
+ \gamma_t \mathbf{1}_t + \beta_6 \text{gen}_z + u_m + \epsilon_{g,m,t}.
\end{align}

\textbf{Lagged Model.} Using generation $g$ predictors to forecast generation $g+1$ outcomes:
\begin{align}
\text{prob\_breakthrough}_{g+1,m,t,z} = &\beta_0^{\text{lag}} +  \nonumber \\ \beta_1^{\text{lag}} H_{\text{fitness},z}(g) + \beta_2^{\text{lag}} H_{\text{spatial},z}(g) \nonumber \\
+ \beta_3^{\text{lag}} \overline{\text{nov}}_z + \beta_4^{\text{lag}} \text{max\_nov}_z \nonumber \\
+ \beta_5^{\text{lag}} (\overline{\text{nov}}_z(g) \times H_{\text{spatial},z}(g)) \nonumber \\
+ \gamma_t^{\text{lag}} \mathbf{1}_t + \beta_6^{\text{lag}} \text{gen}_z(g) \nonumber \\
+ u_m^{\text{lag}} + \epsilon_{g+1,m,t}.
\end{align}

\paragraph{Results Summary:} See Table~\ref{tab:mixedlm_results}

\begin{table*}[htbp]
\centering
\small
\caption{Mixed-Effects Models: Concurrent vs.~Lagged Breakthrough Prediction}
\label{tab:mixedlm_results}
\begin{tabular}{lcccccc}
\toprule
\textbf{Fixed Effect} & \multicolumn{3}{c}{\textbf{Concurrent}} & \multicolumn{3}{c}{\textbf{Lagged (+1 Gen)}} \\
\cmidrule(lr){2-4} \cmidrule(lr){5-7}
& $\hat{\beta}$ & SE & $p$ & $\hat{\beta}$ & SE & $p$ \\
\midrule
\textbf{Main Effects} & & & & & & \\
$H_{\text{fitness},z}$ & $-0.073$ & $0.026$ & 0.005 & $-0.074$ & $0.024$ & 0.002 \\
$H_{\text{spatial},z}$ & $-0.015$ & $0.024$ & 0.532 & $0.012$ & $0.022$ & 0.593 \\
$\text{mean\_novelty}_z$ & $0.070^{**}$ & $0.026$ & 0.006 & $0.016$ & $0.025$ & 0.517 \\
$\text{max\_novelty}_z$ & $0.029$ & $0.023$ & 0.202 & $0.006$ & $0.022$ & 0.787 \\
\textbf{Interaction} & & & & & & \\
$\text{novelty} \times H_{\text{spatial}}$ & $-0.090^{***}$ & $0.010$ & $<0.001$ & $-0.051^{***}$ & $0.009$ & $<0.001$ \\
\textbf{Temporal} & & & & & & \\
$\text{generation}_z$ & $-0.250^{***}$ & $0.018$ & $<0.001$ & $-0.193^{***}$ & $0.017$ & $<0.001$ \\
\midrule
\textbf{Model Statistics} & & & & & & \\
No.~Observations & \multicolumn{3}{c}{3,570} & \multicolumn{3}{c}{3,451} \\
No.~Groups (models) & \multicolumn{3}{c}{15} & \multicolumn{3}{c}{15} \\
Log-Likelihood & \multicolumn{3}{c}{$-4639.77$} & \multicolumn{3}{c}{$-4203.99$} \\
Residual Variance & \multicolumn{3}{c}{$0.7798$} & \multicolumn{3}{c}{$0.6621$} \\
Random Intercept Var & \multicolumn{3}{c}{$0.034$} & \multicolumn{3}{c}{$0.033$} \\
\bottomrule
\end{tabular}
\caption*{\small $^{*}p<0.05$, $^{**}p<0.01$, $^{***}p<0.001$. All predictors z-scored. Task fixed effects included but not separately reported. SE = model-level clustered standard error. Random intercept allows LLM-specific deviation from population mean.}
\end{table*}

\paragraph{Temporal Dynamics:}

The lagged model reveals that current-generation state is a weak predictor of next-generation breakthroughs (residual variance $0.662$ vs.~concurrent $0.780$, suggesting some temporal structure but substantial noise). The interaction term remains the strongest signal across both timescales, suggesting the interference effect is a robust mechanistic feature of LLM-guided evolution, not merely a concurrent correlation.

\section{Supplementary Visualizations}
\label{sec:appendix_additional_results}

\begin{figure}[htbp]
    \centering
    \includegraphics[width=\columnwidth]{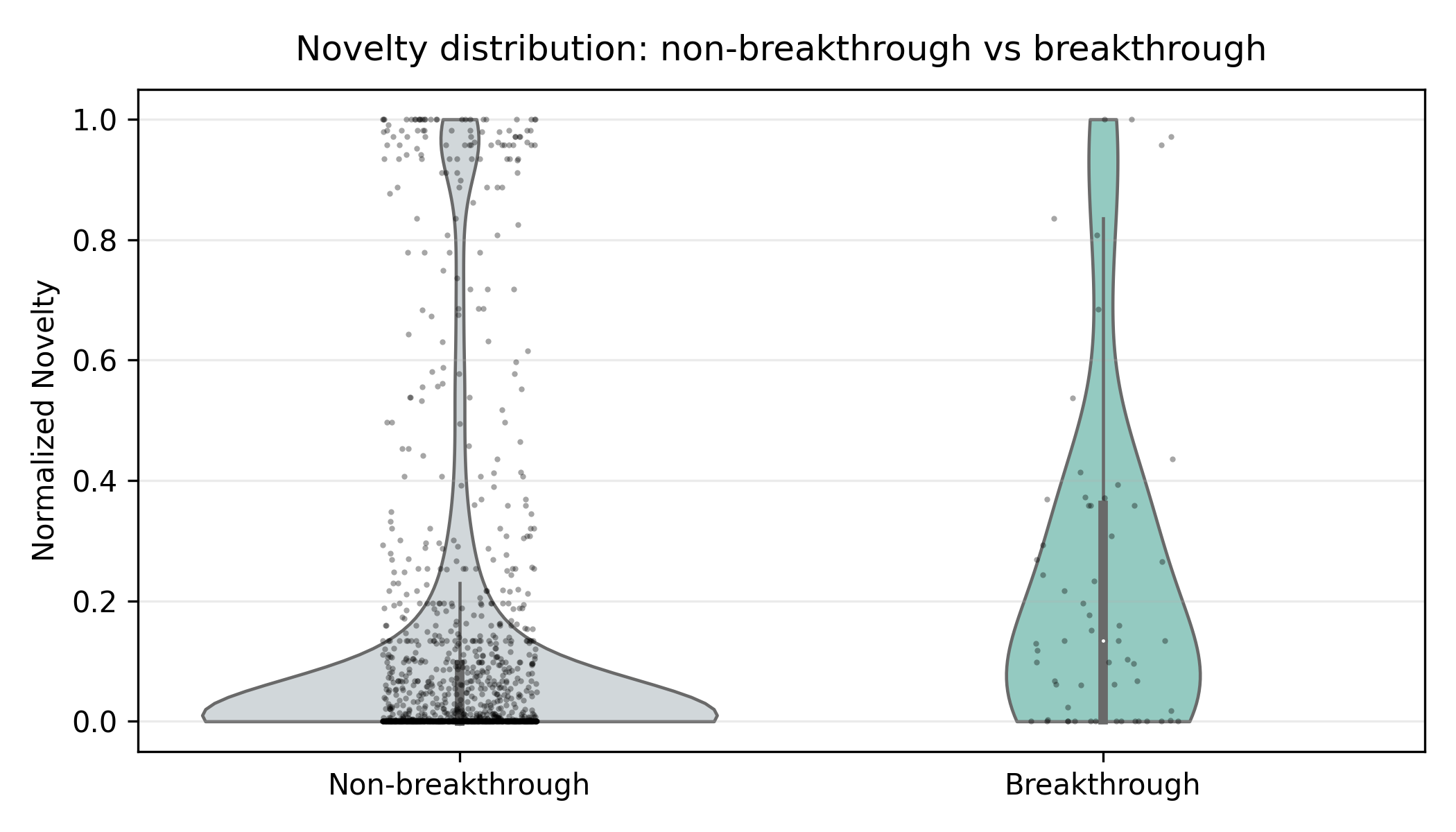}
    \caption{Breakthrough and Non Breakthrough's distribution for all collected trajectories}
    \label{fig:nov_distribution}
\end{figure}

\begin{figure}[htbp]  
    \centering
    \includegraphics[width=\columnwidth]{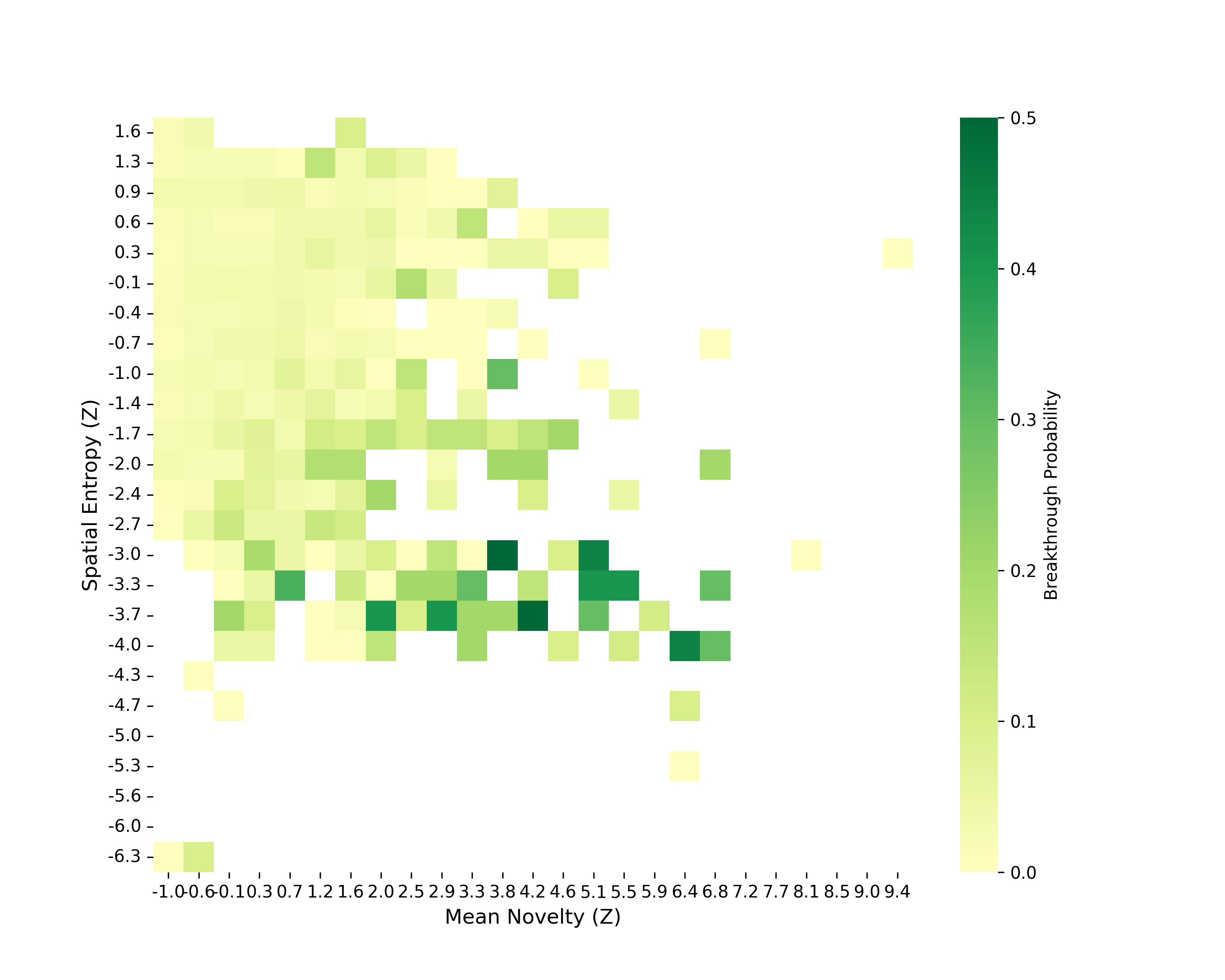}
    \caption{\textbf{Interaction between novelty and spatial entropy in breakthrough dynamics.} Each cell reports the empirical breakthrough probability aggregated over generations falling into the corresponding bins of mean novelty and spatial entropy(z-scored). Color intensity indicates higher likelihood of breakthroughs.}
    \label{fig:interaction_heatmap}
\end{figure}

\begin{figure*}[htbp]
    \centering
    \includegraphics[width=\textwidth]{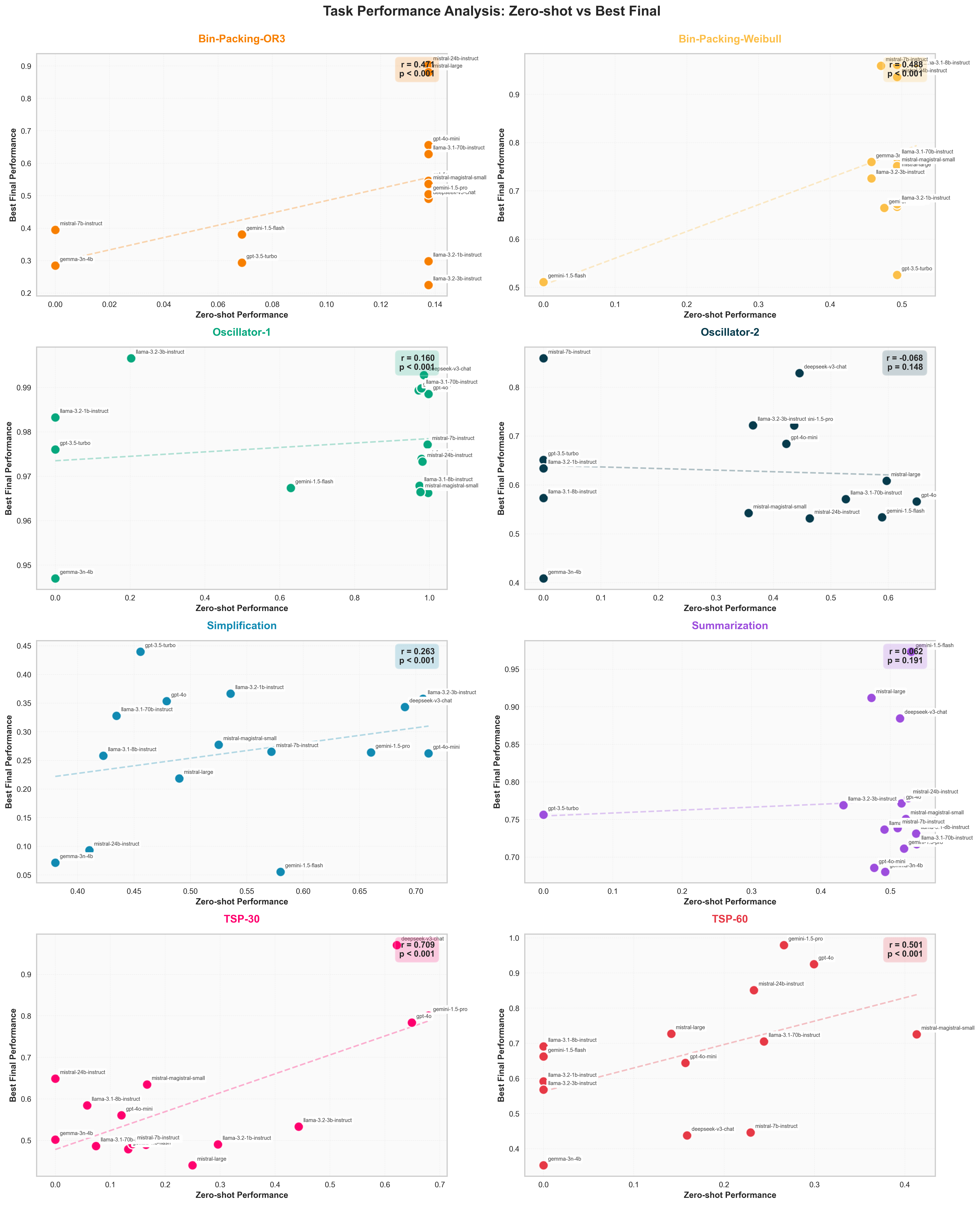}
    \caption{Zero-shot Performance Versus Post-Optimization performance for each task}
    \label{fig:task_zero_perf_best}
\end{figure*}

\begin{figure*}[htbp]
    \centering
    \includegraphics[width=\textwidth]{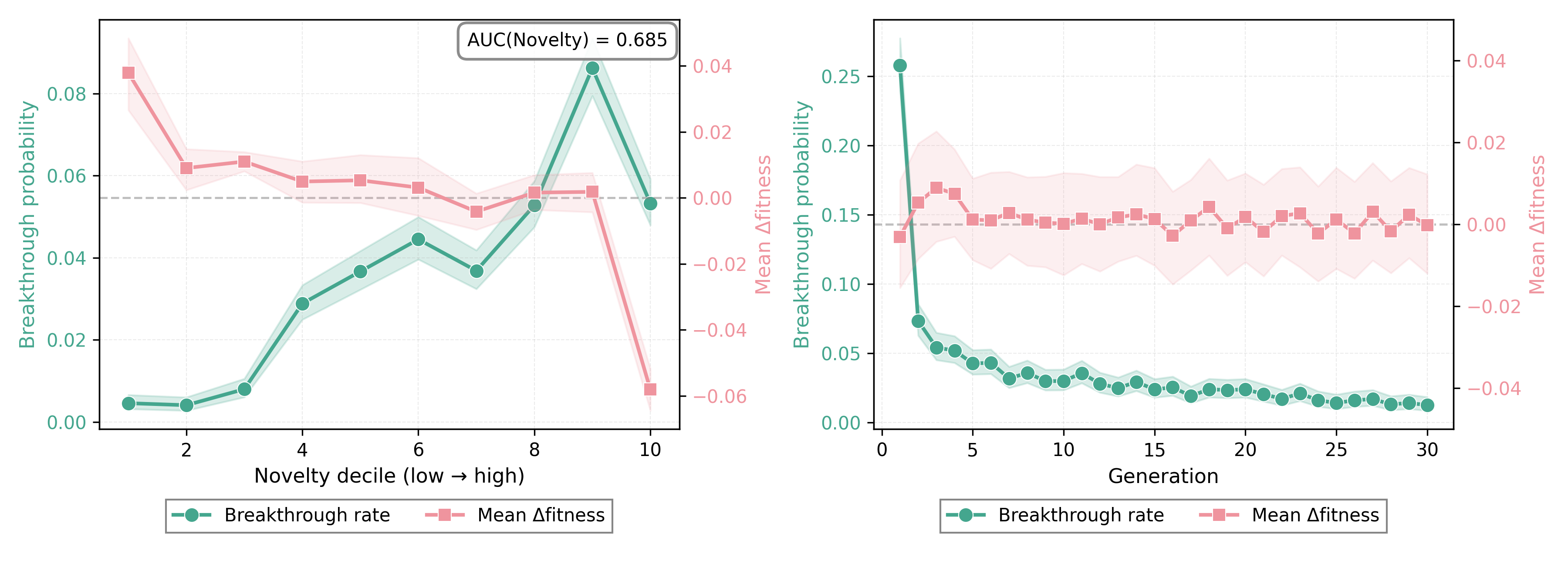}
    \caption{Interaction between breakthroughs and Novelty}
    \label{fig:breakthrough_novelty_analysis}
\end{figure*}

\begin{figure*}[htbp]
    \centering
    \includegraphics[width=\textwidth]{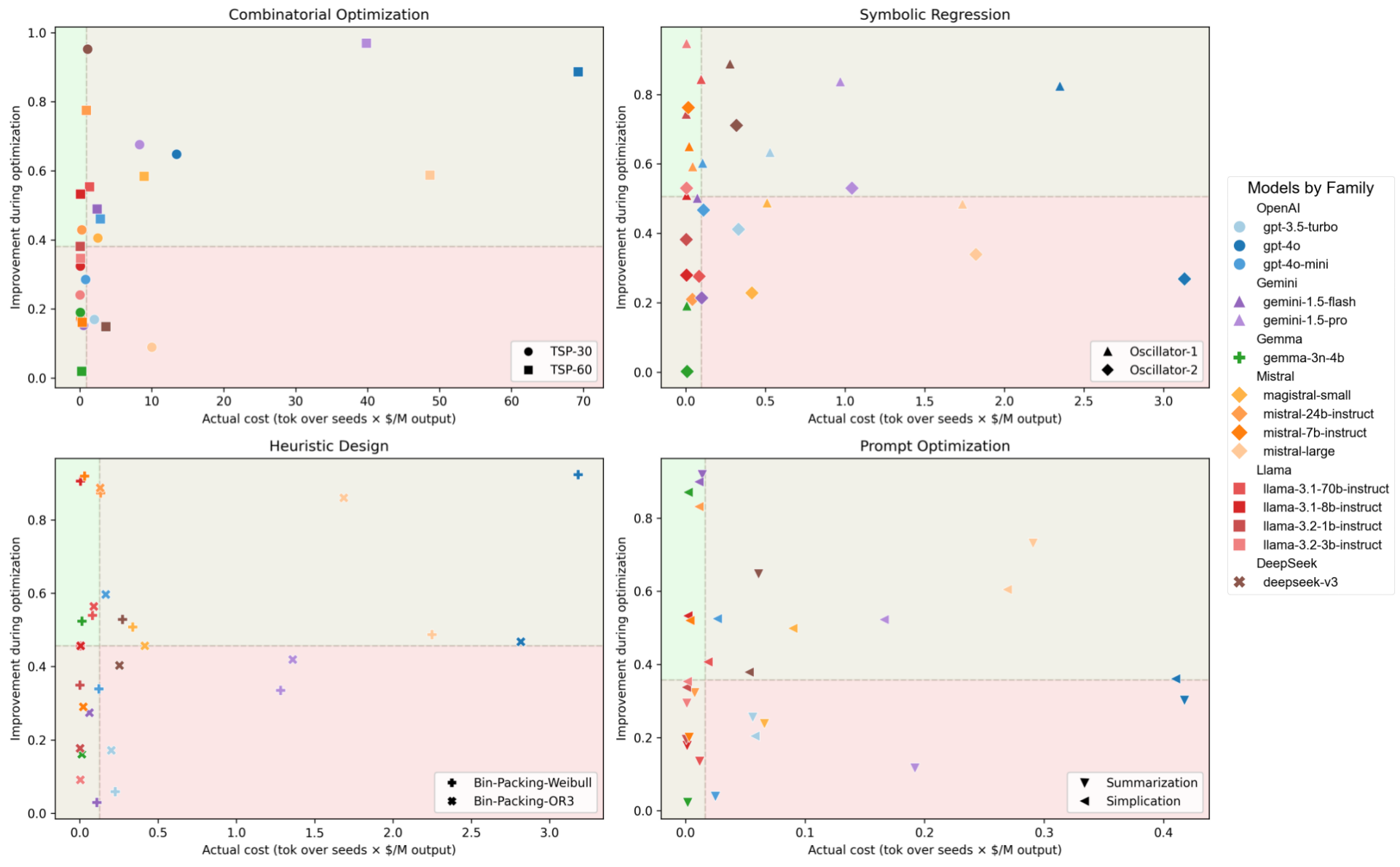}
    \caption{Cost-efficiency plots for four task families}
    \label{fig:cost_four_task}
\end{figure*}

\begin{figure*}[htbp]
    \centering
    \includegraphics[width=\textwidth]{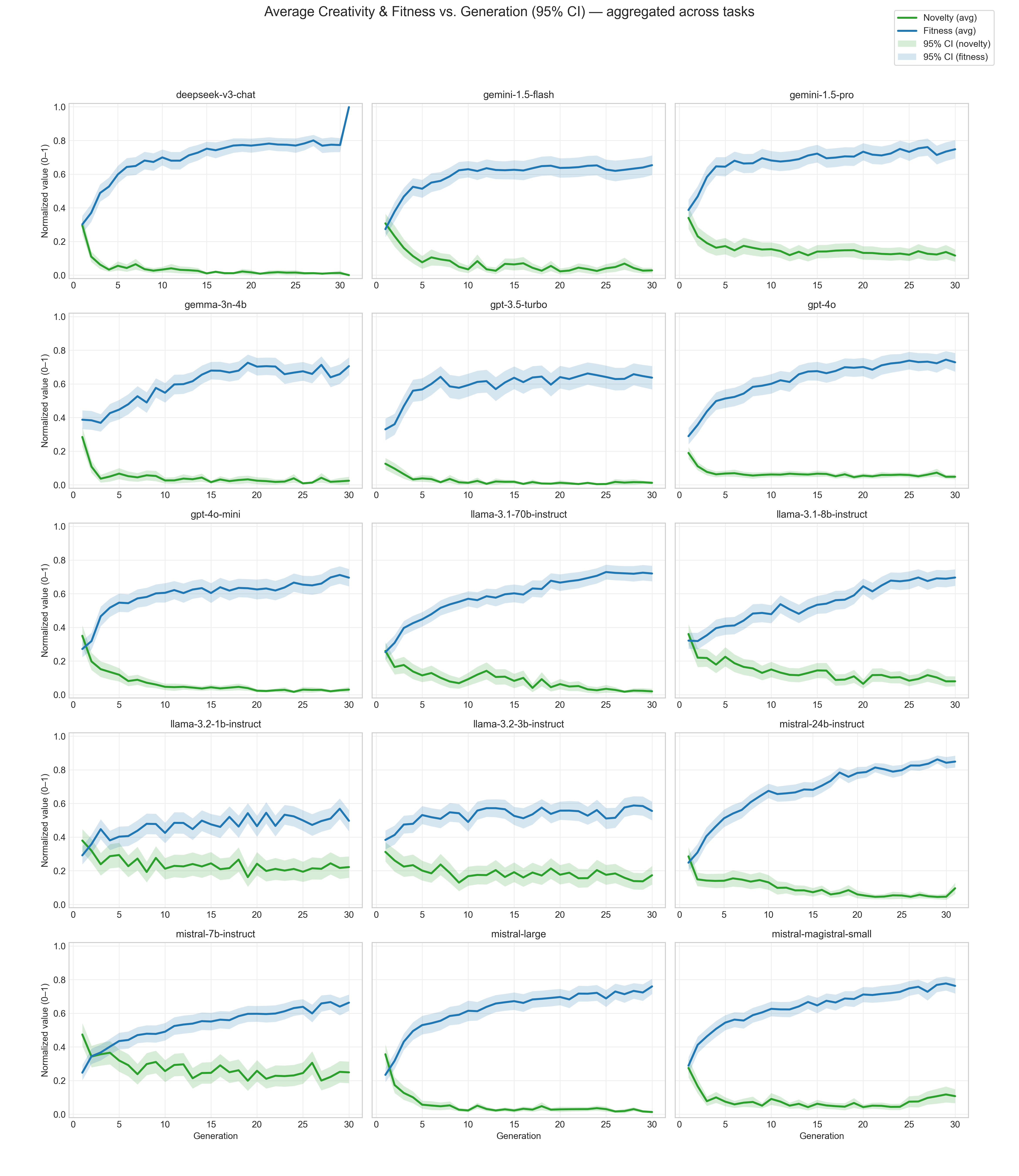}
    \caption{Novelty and fitness coevolution line-plots aggregated over tasks (exploration--exploitation tension)}
    \label{fig:novelty_fitness_coevolve_all_tasks}
\end{figure*}

\end{document}